\documentclass[10pt,twocolumn,twoside]{IEEEtran}

\ifCLASSINFOpdf
\else
\fi
\usepackage{amssymb}
\usepackage{amsmath}
\usepackage{multirow}
\usepackage{bibentry}
\usepackage{color}

\usepackage[pdftex]{graphicx}

\begin{document}
%
\title{Learning to Rank for Blind Image Quality Assessment}
%
%
%

\author{Fei~Gao,~Dacheng~Tao,~\IEEEmembership{Senior Member,~IEEE,}
        Xinbo~Gao,~\IEEEmembership{Senior Member,~IEEE,}
        and~Xuelong~Li,~\IEEEmembership{Fellow,~IEEE}
\thanks{This research was supported partially by the National Natural Science Foundation of China (Grant Nos. 61125106, 61125204, 61432014, and 61172146),
the Key Research Program of the Chinese Academy of Sciences (Grant No. KGZD-EW-T03),
the Fundamental Research Funds for the Central Universities (Grant Nos. K5051202048, BDZ021403 and JB149901),
Microsoft Research Asia Project based Funding (Grant No. FY13-RES-OPP-034),
the Program for Changjiang Scholars and Innovative Research Team in University of China (No. IRT13088),
the Shaanxi Innovative Research Team for Key Science and Technology (No. 2012KCT-02),
and the Australian Research Council Projects (DP-140102164, FT-130101457, and LP-140100569).}
\thanks{Fei Gao is with the VIPS Laboratory, School of Electronic Engineering,
 Xidian University, Xi'an, 710071, Shaanxi, P. R. China. (Email: gaofeihifly@gmail.com). }
\thanks{Dacheng Tao is with the Centre for Quantum Computation \& Intelligent Systems and the Faculty of Engineering and Information Technology,
University of Technology, Sydney, 235 Jones Street, Ultimo, NSW 2007, Australia (Email: dacheng.tao@uts.edu.au).}
\thanks{Xinbo Gao is with the State Key Laboratory of Integrated Services Networks, School of Electronic Engineering,
 Xidian University, Xi'an, 710071, Shaanxi, P. R. China. (Office number: (86) 29 8820-1838 / 2231, Fax number: (86) 29 8820-1620,
Email: xbgao@ieee.org). }
\thanks{Xuelong Li is with the Center for OPTical IMagery Analysis and Learning (OPTIMAL), State Key Laboratory of Transient Optics and Photonics,
Xi'an Institute of Optics and Precision Mechanics, Chinese Academy of Sciences, Xi'an 710119, Shaanxi, P. R. China. (Email: xuelong\_li@opt.ac.cn).}
}

\markboth{IEEE Transactions on Neural Networks and Learning Systems, 2015}%
{Learning to Rank for Blind Image Quality Assessment}
%

\maketitle

\begin{abstract}

Blind image quality assessment (BIQA) aims to predict perceptual image quality scores without access to reference images. State-of-the-art BIQA methods typically require subjects to score a large number of images to train a robust model. However, subjective quality scores are imprecise, biased, and inconsistent, and it is challenging to obtain a large scale database, or to extend existing databases, because of the inconvenience of collecting images, training the subjects, conducting subjective experiments, and realigning human quality evaluations. To combat these limitations, this paper explores and exploits preference image pairs (PIPs) such as ``the quality of image {${\textit{\textbf{I}}}_{\textit{\textbf{a}}}$} is better than that of image {${\textit{\textbf{I}}}_{\textit{\textbf{b}}}$}'' for training a robust BIQA model. The preference label, representing the relative quality of two images, is generally precise and consistent, and is not sensitive to image content, distortion type, or subject identity; such PIPs can be generated at very low cost. The proposed BIQA method is one of learning to rank. We first formulate the problem of learning the mapping from the image features to the preference label as one of classification. In particular, we investigate the utilization of a multiple kernel learning algorithm based on group lasso (MKLGL) to provide a solution. A simple but effective strategy to estimate perceptual image quality scores is then presented. Experiments show that the proposed BIQA method is highly effective and achieves comparable performance to state-of-the-art BIQA algorithms. Moreover, the proposed method can be easily extended to new distortion categories.

\end{abstract}


\section{ INTRODUCTION}

\IEEEPARstart{B}{lind} image quality assessment (BIQA) aims to predict perceptual image quality scores without access to reference images. Because reference images are usually unavailable in most practical applications, BIQA is of great significance and has consequently received tremendous attention over the past decades. To date, a number of universal BIQA methods that work well for various distortion types have been deployed \cite{saad2012BLIINDS-II}-\cite{hou2014DL}. These methods typically require subjects to score a large number of images to learn a robust model, but the acquisition of image quality scores in this way has several limitations.

\begin{figure}[htp]
\vspace{-0.4cm}
\centering
\includegraphics[width=3.5in]{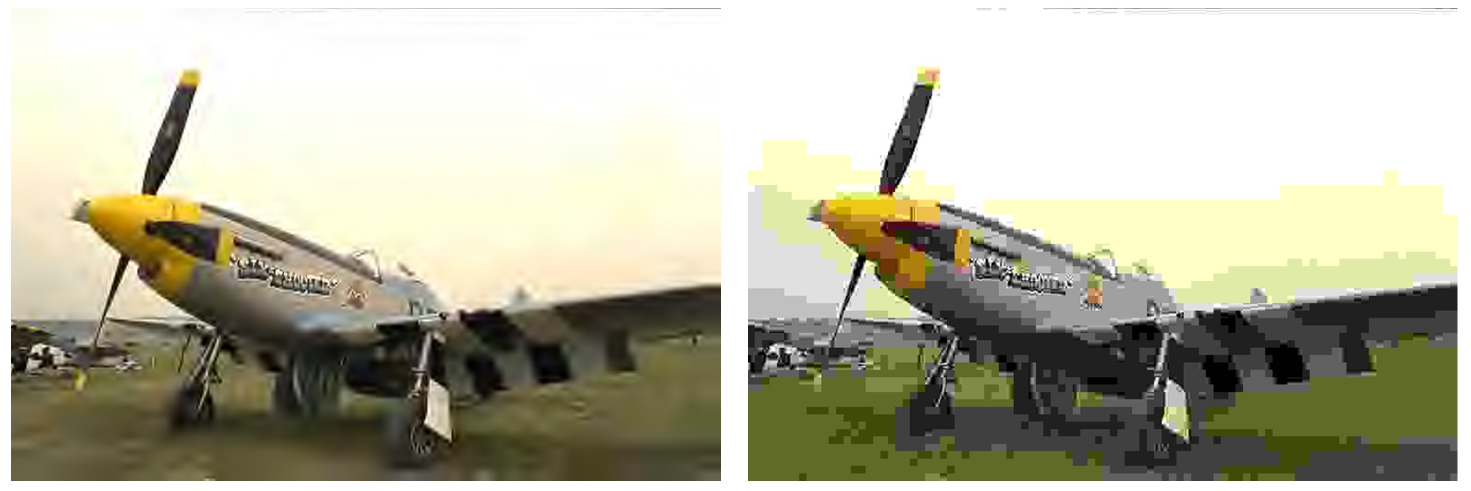}
\vspace{-0.8cm}
\caption{Illustrations of images contained in the LIVE database. Left: JPEG2000 compressed image, $s={\rm 60.77}$, $s_r={\rm 70.84}$; right: JPEG compressed image, $s={\rm 58.63}$, $s_r={\rm 88.92}$.}
\vspace{-0.3cm}
\label{figure1}
\end{figure}

First, scores are not precise. In standard subjective studies \cite{sheikh2006statistical}-\cite{ponomarenko2009tid2008}, the assessor assigns each image a number within a range, e.g. 0 to 100, that reflects its perceived quality. There is usually uncertainty about which score most precisely represents the perceptual quality of a given image \cite{thurstone1994law}. Observers used to hesitantly choose an arbitrary score from the selected range \cite{tsukida2011analyze}. Consequently, this score may not capture subtle differences in the perceived quality of images \cite{rouse2010tradeoffs}. Fig. \ref{figure1} illustrates two images from the Laboratory for Image and Video Engineering (LIVE) database \cite{sheikh2006statistical}. For each image, the difference mean opinion score (DMOS), $s$, and the realigned DMOS, $s_r$, is listed. DMOS is the difference between the mean opinion score (MOS) and perfect quality \cite{sheikh2006statistical}. The realigned DMOS is obtained by realigning the DMOSs between different subjects and between different distortion types. A larger $s$ or $s_r$ value indicates poorer quality. It is notable that the left image is subjectively better than the right one, but the DMOSs do not accurately reflect their relative quality. The success of $s_r$ implies the necessity for realignment.

\begin{figure}[h]
\vspace{-0.4cm}
\centering
\includegraphics[width=3.5in]{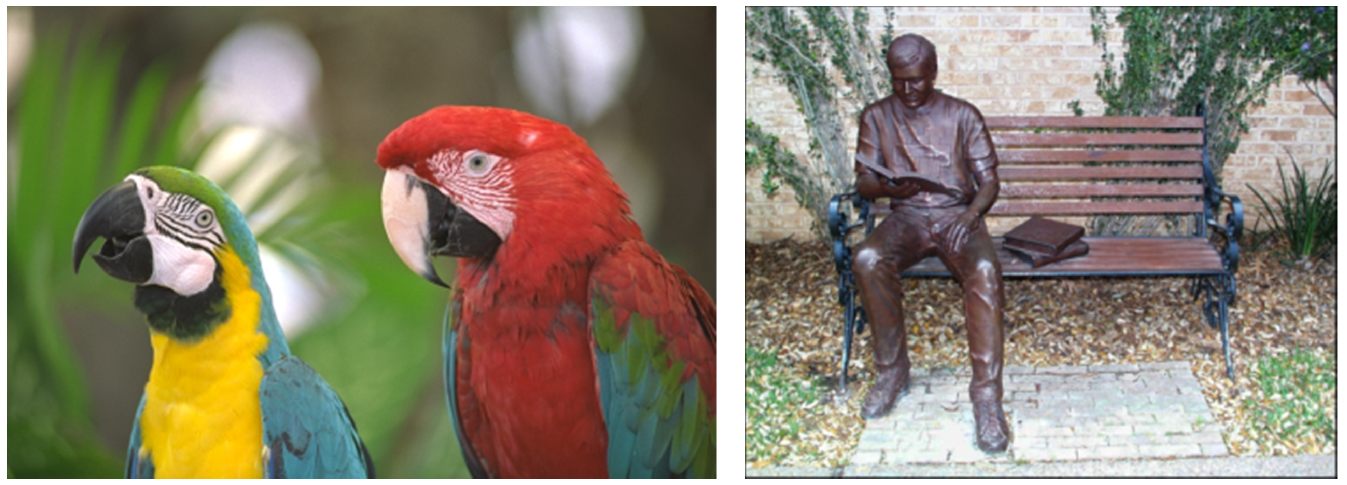}
\vspace{-0.8cm}
\caption{Two undistorted images with different content. Both are of perfect quality.}
\vspace{-0.3cm}
\label{figure2}
\end{figure}

Second, subjective judgments of quality may be biased by image content. Images whose impairment scales are similar but whose content is different may be assigned different scores by observers as a result of personal image content preferences. Fig. \ref{figure2} illustrates two undistorted images, both of which are of perfect quality. However, individuals may prefer the left image and evaluate its quality with a higher score because they find it aesthetically more pleasing than the right image. The content-dependent problem further decreases the reliability of subjective scores.

\begin{figure}[h]
\centering
\vspace{-0.4cm}
\includegraphics[width=3.5in]{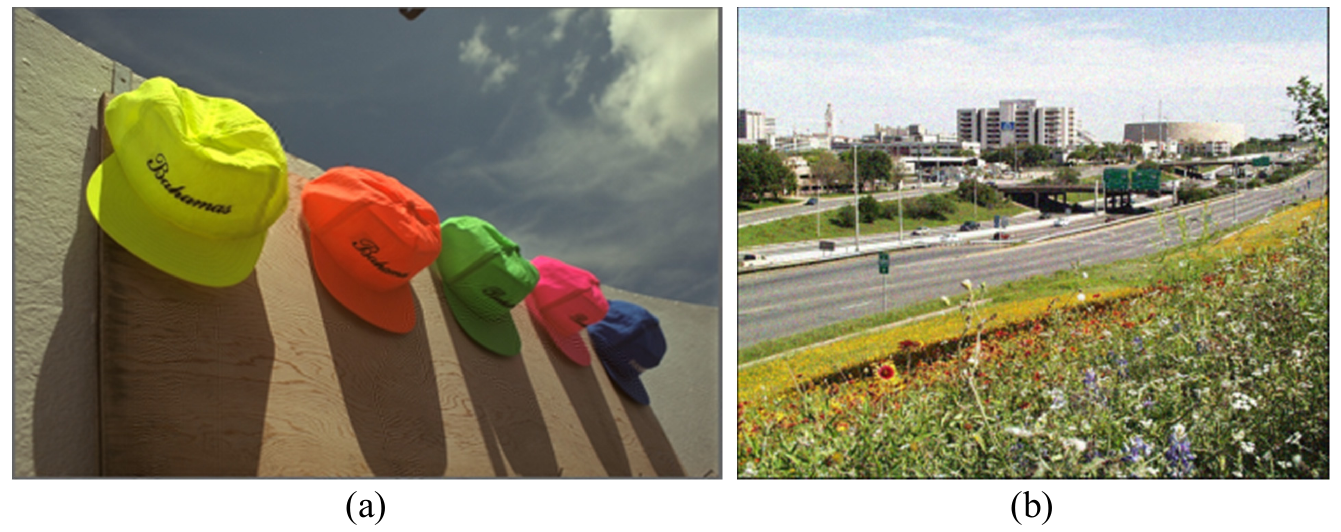}
\vspace{-0.8cm}
\caption{Images corrupted by different distortion types but having similar quality scores: (a) Image distorted by FF,  $s={\rm 43.69}$ and $s_r={\rm 35.42}$; (b) Image corrupted by WN, $s={\rm 32.03}$ and $s_r={\rm 35.32}$.}
\vspace{-0.3cm}
\label{figure3}
\end{figure}

Third, the quality scales between different sessions are inconsistent. In subjective experiments, the evaluation of all the test images is divided into several independent sessions with respect to the distortion type \cite{sheikh2006statistical}\cite{jayaraman2012objective} or image content \cite{TID2013}\cite{ponomarenko2009tid2008} to minimize the effects of observer fatigue \cite{bt2002500}. Thus, images which have similar quality scores but are evaluated in different sessions may not be perceptually similar to each other. Fig. \ref{figure3} shows two images included in the LIVE database. Fig. \ref{figure3}a is an image distorted in a fast fading Rayleigh (wireless) channel (FF), with $s={\rm 43.69}$ and $s_r={\rm 35.42}$. Fig. \ref{figure3}b is an image corrupted by the adaptive Gaussian white noise (WN), with $s={\rm 32.03}$ and $s_r={\rm 35.32}$. Although their realigned DMOSs are almost the same, the impairment in Fig. \ref{figure3}b is much more annoying than it is in Fig. \ref{figure3}a. Again, the raw DMOS fails to reflect the relative quality.

Finally, it is challenging to obtain a large scale database or to extend existing databases, mainly for the following inconvenient reasons: 1) The test organizer has to collect sufficient images associated with each kind of distortion at diverse levels of degradation to yield a meaningful evaluation \cite{bt2002500}; 2) One observer has to continuously evaluate many images in a single session to minimize the influence of contextual effects that the score to an image is highly biased by recently scored images \cite{de1996current}\cite{parducci1986category}; 3) Subjective experiments should be conducted in critical viewing conditions and in critical procedures, because judgments of perceptual quality scores are sensitive to the viewing environment \cite{rouse2010tradeoffs}\cite{bt2002500}; 4) It is necessary to recruit many observers and train them before the experiment, to minimize quality scale mismatch errors \cite{sheikh2006statistical}; and 5) The process of realigning raw human responses is complicated \cite{sheikh2006statistical}.
\begin{figure}[h]
\centering
\vspace{-0.4cm}
\includegraphics[width=3.5in]{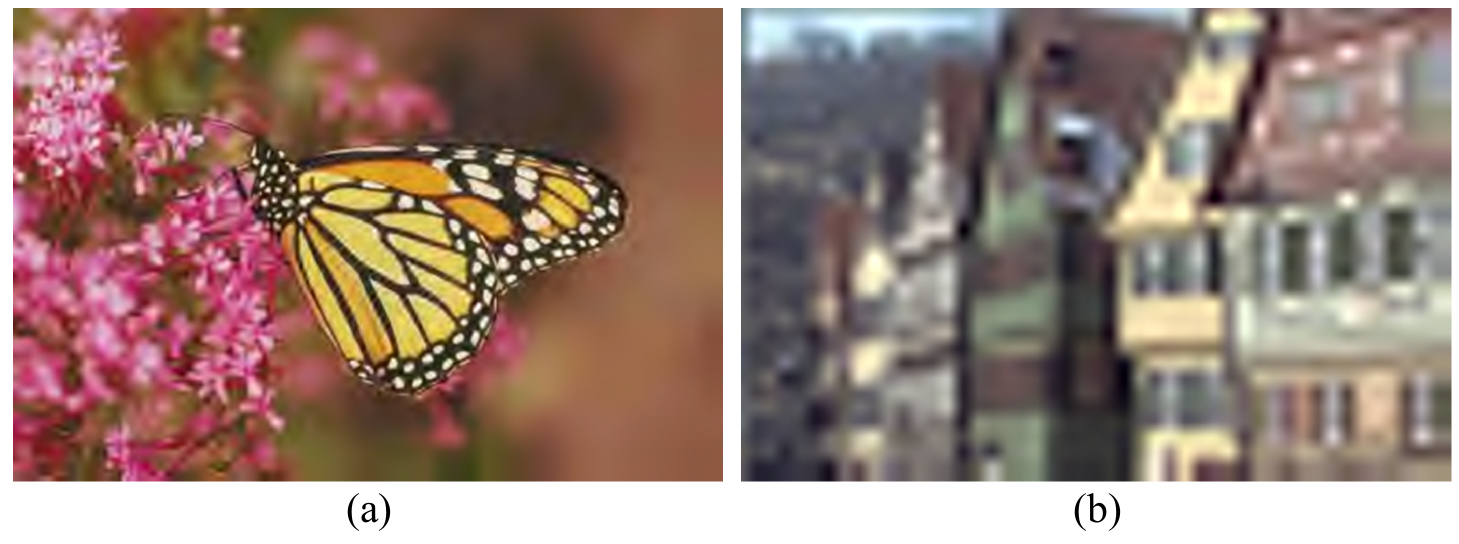}
\vspace{-0.8cm}
\caption{Images diverse in image content, distortion type, and distortion strength: (a) JPEG2000 compressed image, (b) image distorted by FF.}
\vspace{-0.3cm}
\label{figure4}
\end{figure}

These limitations constrain the reliability and extension of existing BIQA methods which utilize subjective quality scores as the benchmark for learning a model. Although there have been several BIQA methods that do not require learning from subjectively evaluated images [23]-[25], their performance is still inferior to state-of-the-art algorithms.

To combat the aforementioned limitations, this paper explores and exploits preference image pairs (PIPs) such as "the quality of image $I_a$ is better than that of image $I_b$" for training a robust BIQA model. The preference label, representing the relative quality of two images, is generally precise and consistent, and is not sensitive to image content, distortion type, viewing conditions, or subject identity \cite{rouse2010tradeoffs}\cite{bradley1952rank}, and such PIPs can be generated at very low cost \cite{bradley1952rank}. Consider the image pairs derived from the images shown in Figs. \ref{figure3} and \ref{figure4}. They are easy to discriminate in terms of perceived quality. Moreover, different subjects may give consistent preference labels \cite{halonen2011naturalness}, even though the images in a certain pair may vary in the types of distortion afflicting them (e.g. Figs. \ref{figure4}a and \ref{figure4}b), or in image content (e.g. Figs. \ref{figure3}a and \ref{figure4}b), or both (e.g Figs. \ref{figure3}b and \ref{figure4}b).

The proposed BIQA method is one of learning to rank \cite{hang2011short}. We first formulate the problem of learning the mapping from the image features to the preference label as one of classification. In particular, we investigate using a multiple kernel learning algorithm based on group lasso (MKLGL) \cite{xu2010simple} to solve it. A simple but effective strategy is then presented to estimate perceptual image quality scores. Thorough experiments conducted on the largest four standard databases show that the proposed BIQA method is highly effective and achieves comparable performance to state-of-the-art BIQA algorithms. Moreover, the proposed method can be easily extended to new distortion categories by simply adding the corresponding PIPs into the training set.

The rest of the paper is organized as follows. In Section \ref{sec:Related Work}, we briefly introduce standard IQA databases and learning to rank. Section \ref{sec:Generation of Preference Image Pairs} discusses how to generate PIPs with valid preference labels. The framework of the proposed BIQA method is detailed in Section \ref{sec:The Proposed BIQA Method}. Our experiments are described and analyzed in Section \ref{sec:IQA Experiments}, and an extensive subjective study of PIPs is presented in Section \ref{Subjective Study PIPs}. Section \ref{sec:Conclusion} concludes this paper.

\vspace{-0.3cm}\section{ Related Work}
\label{sec:Related Work}

\subsection{Existing IQA Databases}
\label{sec:Existing IQA Databases}

Subjective IQA studies are of fundamental importance for the development of IQA. Over the years, many researchers have contributed significant research in this area through the construction of various IQA databases. In this section, we introduce the composition of the four largest datasets, i.e. the LIVE dataset \cite{sheikh2005live}, Tampere Image Database 2013 (TID2013) \cite{TID2013}, Categorical Subjective Image Quality (CSIQ) database \cite{larson2010CSIQ}, and LIVE multiple distorted (LIVEMD) database \cite{jayaraman2012objective}.

The LIVE database includes 808 images, which are generated from 29 original images by corrupting them with five types of distortion, i.e. JPEG2000 compression (JP2k), JPEG compression (JPEG), WN, Gaussian blur (Gblur), and FF. The DMOS and realigned DMOS of each image are available. Because the realigned DMOS is more precise than the DMOS, we adopt the realigned DMOS in our work. The realigned DMOS ranges from -3 to 112. For convenience, we refer to the realigned DMOS as DMOS in the remainder of this paper unless otherwise indicated. In addition, we refer to all the images associated with the same reference image as a ``group''.

The TID2013 includes 3,000 images in sum. These images are generated by corrupting 25 original images with 24 types of distortion at 5 different levels. The distortion types include: WN (\#1), additive white Gaussian noise which is more intensive in color components than in the luminance component (\#2), additive Gaussian spatially correlated noise (\#3), masked noise (\#4), high frequency noise (\#5), impulse noise (\#6), quantization noise (\#7), Gblur (\#8), image denoising (residual noise, \#9), JPEG (\#10), JP2k (\#11), JPEG transmission errors (\#12), JPEG2000 transmission errors (\#13), non-eccentricity pattern noise (\#14), local block-wise distortion of different intensity (\#15), mean shift (\#16), contrast change (\#17), change of color saturation (\#18), multiplicative Gaussian noise (\#19), comfort noise (\#20), lossy compression of noisy images (\#21), image color quantization with dither (\#22), chromatic aberrations (\#23), and sparse sampling and reconstruction (\#24). The MOS of each image is available and ranges from 0.2 to 7.3.

The CSIQ database consists of 866 images which are derived from 30 original images. Six types of distortion are considered in CSIQ: WN, JPEG, JP2k, additive Gaussian pink noise (PN), Gblur, and global contrast decrements (GCD). The DMOS of each image is available and ranges from 0 to 1.

The LIVEMD database includes images distorted by multiple types of distortion. There are two subsets, one of which is associated with the images corrupted by Gblur followed by JPEG (GblurJPEG), and one which is associated with images corrupted by Gblur followed by WN (GblurWN). Each subset includes 225 images. The DMOS of each image is released and ranges from 0 to 85.

\vspace{-0.3cm}
\subsection{Learning to Rank}
\label{sec:Learning to Rank}
Learning to rank \cite{hang2011short} involves learning a function that can predict the ranking list of a given set of stimuli. It is the central issue in web page ranking, document retrieval, image searching and other applications \cite{Li2014rank1}\cite{Li2014rank2}. Of the existing methods \cite{Li2014rank3}, the pairwise approach has been well deployed and successfully applied to information retrieval \cite{cao2007learning}\cite{Leonardo2011sortnet}.

Without loss of generality, we take document retrieval as an example to introduce the pairwise approach. In pairwise approaches, document pairs are adopted as instances for learning a ranking function. What is needed is to construct a training set by first collecting document pairs from the ranking lists, and then calculating the corresponding difference feature vector for each pair of documents and assigning a preference label that represents their relative relevance \cite{chapelle2010efficient}. In this way, the problem of learning-to-rank is reduced to a classification problem, also called \textit{preference learning} problem \cite{hullermeier2008label}, and existing classification models, such as support vector machines (SVMs) and Neural Network, can be directly applied to solve it \cite{herbrich1999large}\cite{zheng2007general}.

Intuitively, the problem addressed in the pairwise approach is similar to the problem we aim to work out in this paper, and is our inspiration for thinking we may succeed in learning a robust BIQA model from PIPs. The proposed BIQA model is detailed in Section \ref{sec:The Proposed BIQA Method}.

\vspace{-0.3cm}
\section{Generation of Preference Image Pairs}
\label{sec:Generation of Preference Image Pairs}
Generating PIPs in an easy and efficient way is of fundamental significance in this research. In this section, we investigate how to produce image pairs with valid preference labels through paired comparisons and from existing IQA databases, based on the quality scores.

\vspace{-0.32cm}
\subsection{Paired Comparisons}
\label{sec:Paired Comparisons}
In paired comparisons, observers are asked to choose which image in a pair has better perceived quality \cite{bradley1952rank}. In the scenario of this paper, image quality is defined as the fidelity of a distorted image comparing to the corresponding undistorted image. Given an image, observers assess its quality based on the impairments contained in it \cite{bt2002500}. This definition is consistent with the majority of existing literature about image quality assessment, e.g. \cite{sheikh2005live}-\cite{ponomarenko2009tid2008},\cite{bt2002500}. Consequently, in paired comparisons, observers should report the preference label of an image pair by simply comparing the impairments in the images. As a result, we can combat the content preference problem in most cases.

Consider a pair of images. If the difference in impairment scale between them is sufficiently large, observers can easily discriminate between them, regardless of any type of corruption through distortion, image content, or even viewing conditions. Moreover, different observers may offer a unanimous judgment \cite{halonen2011naturalness}. In this case, we can obtain valid preference labels by arranging only one observer to assign each pair. This is consistent with the samples illustrated in Figs. \ref{figure3} and \ref{figure4}.

\begin{figure}[h]
\centering
\vspace{-0.4cm}
\includegraphics[width=3.5in]{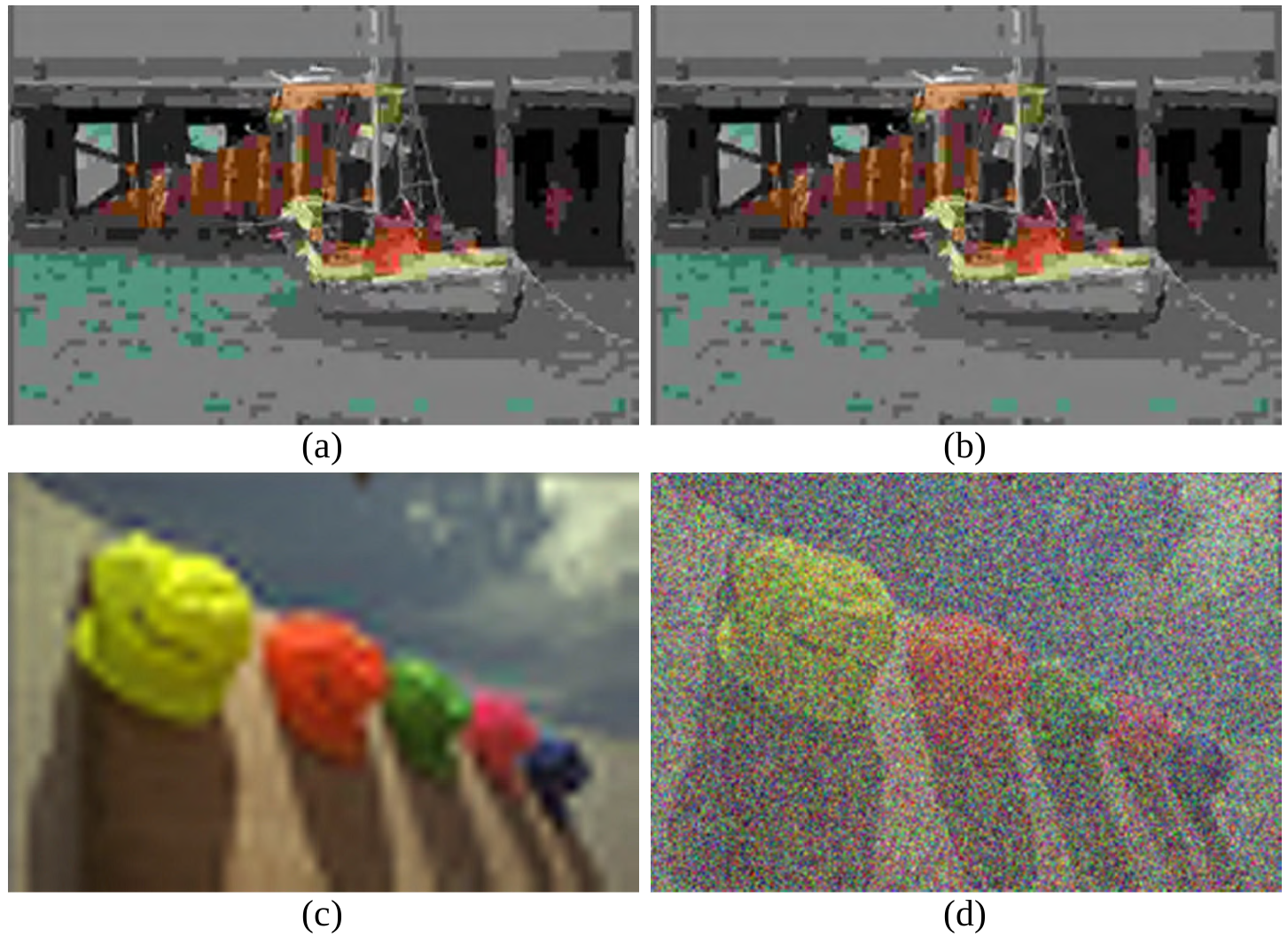}
\vspace{-0.8cm}
\caption{Distorted images: (a) JPEG compressed image; (b) JPEG compressed image; (c) image corrupted by FF; (d) image corrupted by WN.}
\vspace{-0.32cm}
\label{figure5}
\end{figure}

However, when the distortion strengths of the two images are similar, it becomes much more difficult for observers to judge their relative quality, especially when the image content of the two images is diverse (e.g. Figs. \ref{figure5}b and \ref{figure5}c), or because of the distortion type (e.g. Figs. \ref{figure5}c and \ref{figure5}d), or both (e.g. Figs. \ref{figure5}b and \ref{figure5}d), or because there is subtle difference between them (e.g. Figs. \ref{figure5}a and \ref{figure5}b). In this case, it is unreasonable neither to force observers to choose a ``better'' image \cite{TID2013}\cite{jayaraman2012objective}, nor to label the relative quality as ``the same'' \cite{bt2002500}. Otherwise, discrepancies are caused and responses from many observers are needed \cite{rouse2010tradeoffs} in order to produce valid preference labels. The huge number of comparisons coupled with the need to recruit many observers \cite{TID2013}\cite{jayaraman2012objective} lead to the same limitations as introduced previously.

Thus we propose to allow observers to only judge the image pairs that are easy to distinguish in terms of perceived quality in paired comparisons. Since we aim to learn a BIQA model from image pairs, we do not require a complete comparison, but only to label a sample of all possible pairs. We conclude the procedures of generating PIPs through paired comparisons as follows:

First, collect $n$ images, ${\mathcal I}{\rm =}{\rm \{}I_1{\rm ,\ ...,\ }I_n\}{\rm ,\ }n>2$, are diverse in the types of corrupting distortion, image content, and distortion strength.

Second, randomly construct $N{\rm ,\ 1}\le N\le n(n-1)/2$, pairs from the collected images. The image pair set ${\mathcal P}$ is given by
\begin{equation}
{\mathcal P}\subseteq \left\{\left(I_i,I_j\right)\mathrel{\left|\vphantom{\left(I_i,I_j\right) i,j=1,2,\cdots ,n}\right.\kern-\nulldelimiterspace}i,j=1,2,\cdots ,n\right\}.
\end{equation}
\noindent Note that if $(I_i,I_j)\in {\mathcal P}$, then $(I_j,I_i)\notin {\mathcal P}$, to reduce the number of comparisons.

Finally, recruit $S,\ S\ge 1$, subjects to assign preference labels for the image pairs. Given a pair $(I_i,I_j)$, if subject $s,\ s=1,\dots ,S$, considers $I_i$ to be better than $I_j$ in quality, then the corresponding preference label $l_{k,s}$, $k=1,2,\cdots ,N$, is set as $+1$; if $I_j$ is subjectively better than $I_i$, $l_{k,s}=-1$; and if subject $s$ does not label the pair $(I_i,I_j)$ or is uncertain about the relative quality of the corresponding images, let $l_{k,s}=0$. For simplicity, the final preference label for the pair $(I_i,I_j)$ is calculated as
\begin{equation}
y_k={\rm sign}\left({\sum}_{s=1}^S{l_{k,s}}\right).
\end{equation}
\noindent If $y_k=0$, it means $I_i$ and $I_j$ are difficult to discriminate in terms of perceived quality, and $(I_i,I_j)$ will not be adopted for training.

The procedure of the paired comparison experiment is flexible \cite{bradley1952rank}. Observers do not have to evaluate many pairs in a single session, because there is practically no contextual effect or scale mismatch problem, and PIPs evaluated in different sessions can be directly aggregated into a single dataset for future applications without complex processing. In this paper, we present an extensive subjective paired-comparison study which we detail in Section \ref{Subjective Study PIPs}.

\vspace{-0.3cm}
\subsection{ Collecting PIPs from Existing IQA Databases }
\label{sec:Collecting PIPs from Existing IQA Databases }

Since many efforts have contributed to the construction of various IQA databases \cite{sheikh2005live}-\cite{jayaraman2012objective}, it is meaningful to find a method to generate reliable PIPs from existing databases. Previous discussions imply that a small difference between two quality scores may not accurately reflect the corresponding relative quality; thus, we recommend selecting the image pairs whose corresponding difference quality scores reach a threshold. The generation of PIPs from IQA databases is therefore formulated as below.

Assume that an IQA database comprises: 1) \textit{n }labeled images: ${\mathcal I}{\rm =}{\rm \{}I_1{\rm ,\ ...,\ }I_n\}$; 2) the subjective quality scores of the images ${\mathcal Q}{\rm =}{\rm \{}q_1{\rm ,\ ...,\ }q_n\}$, $q_i$ is the quality score of $I_i$. Then we can construct $N$ image pairs from these labeled images. Let ${\mathcal P}$ be the pair set, we have
\begin{equation}
{\mathcal P}\subseteq \left\{\left(I_i,I_j\right)\mathrel{\left|\vphantom{\left(I_i,I_j\right) \ \left|q_i-q_j\right|>T\ \ i,j=1,\dots ,n}\right.\kern-\nulldelimiterspace}\ \left|q_i-q_j\right|>T\ \ i,j=1,\dots ,n\right\},
\end{equation}
\noindent where, $T$ is the threshold of the difference quality score, and $T\ge 0$; and$\ \left|q_i-q_j\right|$ denotes the absolute value of $\left(q_i-q_j\right)$. Because a greater MOS value indicates better quality but a greater DMOS value is associated with poorer quality, a preference label $y_k,\ k=1,\dots ,N$ for each image pair $\left(I_i,I_j\right)\in {\mathcal P}$\textit{ }is assigned as:
\begin{equation}
y_k{\rm =}\left\{ \begin{IEEEeqnarraybox}[\relax][c]{l's}
{\rm sign}\left(q_i-q_j\right), & if the quality score is MOS \\
{\rm -}{\rm sign}\left(q_i-q_j\right), & if the quality score is DMOS \end{IEEEeqnarraybox}\right.
\end{equation}
Thus, $y_k=1$ indicates that $I_i$ is better than $I_j$ in terms of quality, but $y_k=-1$ indicates that $I_j$ is better. Let ${\mathcal Y}$ be the set of preference labels, such that
\begin{equation}
{\mathcal Y}{\mathbf =}\left\{y_1,\dots ,y_N\right\}\subset {\left\{{\rm -}{\rm 1,+1}\right\}}^N.
\end{equation}

It is notable that the PIPs separately extracted from different IQA databases can be combined with no realignment. Moreover, we can aggregate the PIPs produced through paired comparisons and those collected from IQA databases into one single dataset for training a BIQA model. Based on all the previous discussions, we conclude that the generation of PIPs is easy and convenient, avoiding the limitations in the acquisition of human quality scores.

\vspace{-0.3cm}
\section{ The Proposed BIQA Method}
\label{sec:The Proposed BIQA Method}

In this section, we present a BIQA framework that learns to predict perceptual quality scores from PIPs. Inspired by the pairwise learning-to-rank approaches, we first formulate the problem of learning the mapping from difference feature vectors to preference labels as one of classification. We utilize natural scene statistics (NSS) \cite{srivastava2003advances}-based features to represent an image and investigate using the MKLGL approach \cite{xu2010simple} to solve the classification problem. A simple but effective strategy is subsequently presented to estimate perceptual image quality scores. Fig. 6 shows the diagram of the proposed BIQA method, and details are introduced in the following subsection.

\begin{figure*}[ht]
\centering
\vspace{-0.4cm}
\includegraphics[width=6in]{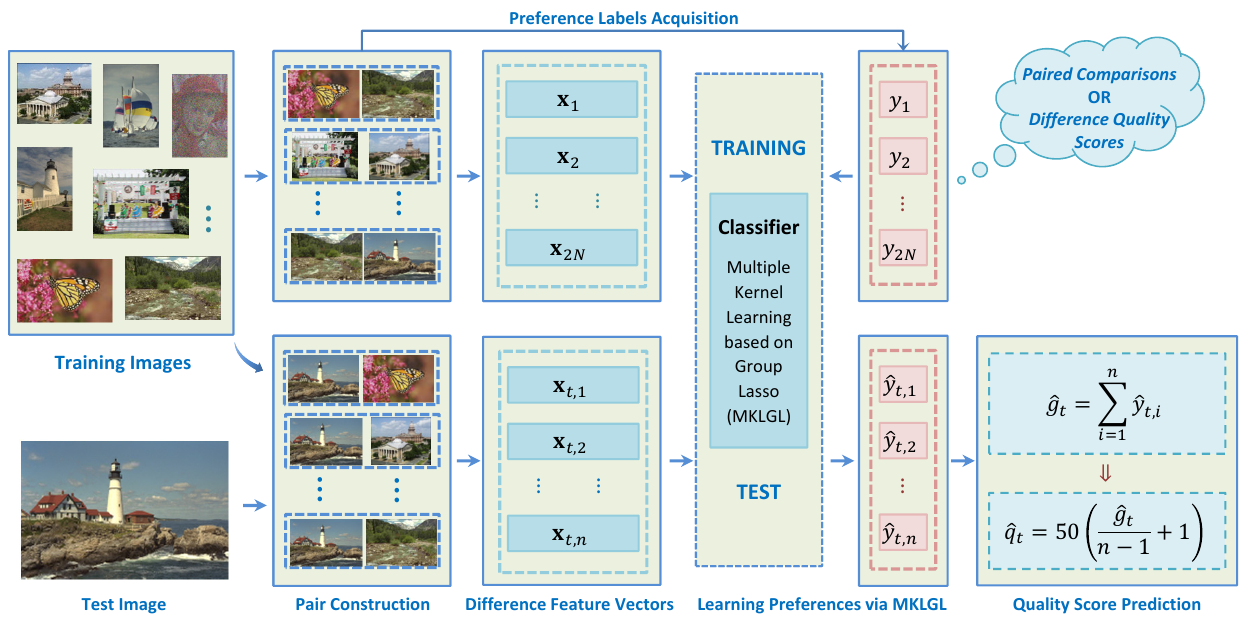}
\vspace{-0.3cm}
\label{figure6}
\caption{Diagram of the proposed BIQA method.}
\vspace{-0.3cm}
\end{figure*}

\vspace{-0.3cm}
\subsection{Integrated NSS Features}
\label{sec:Integrated NSS Features}

NSS-based image features have been widely explored for IQA and have shown promising performance \cite{saad2012BLIINDS-II}-\cite{he2012SRNSS}. However, the features utilized in these methods are generally representative of some distortion categories but have relatively weaker correlation with others. In this paper, therefore, we adopt a fusion of the features that have been utilized in several state-of-the-art BIQA methods, i.e. BLIINDS-II \cite{saad2012BLIINDS-II}, BRISQUE-L \cite{mittal2012BRISQUE-L}, and SRNSS \cite{he2012SRNSS}, to represent an image. For the integrity of this paper, we briefly introduce these features in this subsection. For detail, we refer to \cite{saad2012BLIINDS-II}, \cite{mittal2012BRISQUE-L}, and \cite{he2012SRNSS}.

Both the BLIINDS-II features and BRISQUE-L features are extracted over several scales, where the feature extraction is repeated after downsampling it by a factor of 2. BLIINDS-II extracts features over 3 scales and computes 8 features in the discrete cosine transform (DCT) domain at each scale. The BRISQUE-L features are exacted over 2 scales, and 18 spatial-domain features are extracted at each scale. The BLIINDS-II and BRISQUE-L features extracted at each scale are concentrated into a vector and denoted as ${{\mathbf f}}^{BLD}_n,\ n=1,2,3$, and ${{\mathbf f}}^{BRL}_m$, $m=1,\ 2$, respectively. In addition, let ${{\mathbf f}}^{BLD}\ $and ${{\mathbf f}}^{BRL}$ be all the BLIINDS-II and BRISQUE-L features, respectively.

In SRNSS, He \textit{et al.} \cite{he2012SRNSS} employed the mean, variance, and entropy of wavelet coefficients in each sub-band to encode the generalized spectral behavior, the energy fluctuations, and the generalized information, respectively. In the implementation, an image is decomposed into 4 scales. The coefficients at LH (low, high) and HL (high, low) sub-bands at a particular scale are combined to calculate a group of features due to their similarity in statistics. Let ${\mathbf m}$, ${\mathbf v}$, and ${\mathbf e}$ respectively denote all the means, variances, and entropies extracted from an image, and ${{\mathbf f}}^{SRN}$ be all the SRNSS features extracted from an image.

Finally we denote all the features extracted from an image as ${{\mathbf f}}^{all}$, such that
\begin{eqnarray}
\begin{split}
{{\mathbf f}}^{all} & = \left[{{\mathbf f}}^{BLD},\ {{\mathbf f}}^{BRL},\ {{\mathbf f}}^{SRN}\right] \\
& =\left[{{\mathbf f}}^{BLD}_1,{{\mathbf f}}^{BLD}_2,{{\mathbf f}}^{BLD}_3,{\mathbf \ }{{\mathbf f}}^{BRL}_1,{{\mathbf f}}^{BRL}_2,{\mathbf m},{\mathbf \ }{\mathbf v},{\mathbf \ }{\mathbf e}\right].
\end{split}
\end{eqnarray}
\noindent Thus ${{\mathbf f}}^{all}$ is of the dimension 84 (8$\times $3 of ${{\mathbf f}}^{BLD}$+ 18$\times $2 of ${{\mathbf f}}^{BRL}$ + 8$\times $3 of ${{\mathbf f}}^{SRN}$).

We calculated \textit{Pearson's linear correlation coefficient} (PLCC) between each of these features and the quality score, and plotted the maximum PLCC values of ${{\mathbf f}}^{BLD}\ $, ${{\mathbf f}}^{BRL}$, and ${{\mathbf f}}^{SRN}$, respectively, across all the images contained in each type of distortion and the whole image set ({\small \sf ALL}) in the LIVE, CSIQ, TID2013, and LIVEMD databases in Fig. 7. As is clear, ${{\mathbf f}}^{BLD}$, ${{\mathbf f}}^{BRL}$, and ${{\mathbf f}}^{SRN}$ do not correlate highly with all the distortion categories. It is therefore reasonable to combine these features as the representation of the image \cite{lanckriet2004statistical}. In Section \ref{sec:IQA Experiments}, we compare the performance of the BIQA methods that utilize ${{\mathbf f}}^{BLD}$, ${{\mathbf f}}^{BRL}$, ${{\mathbf f}}^{SRN}$, and ${{\mathbf f}}^{all}$, respectively, to verify the effectiveness of this feature fusion strategy.

    \begin{figure*}[t]
    \centering
    \includegraphics[width=6in]{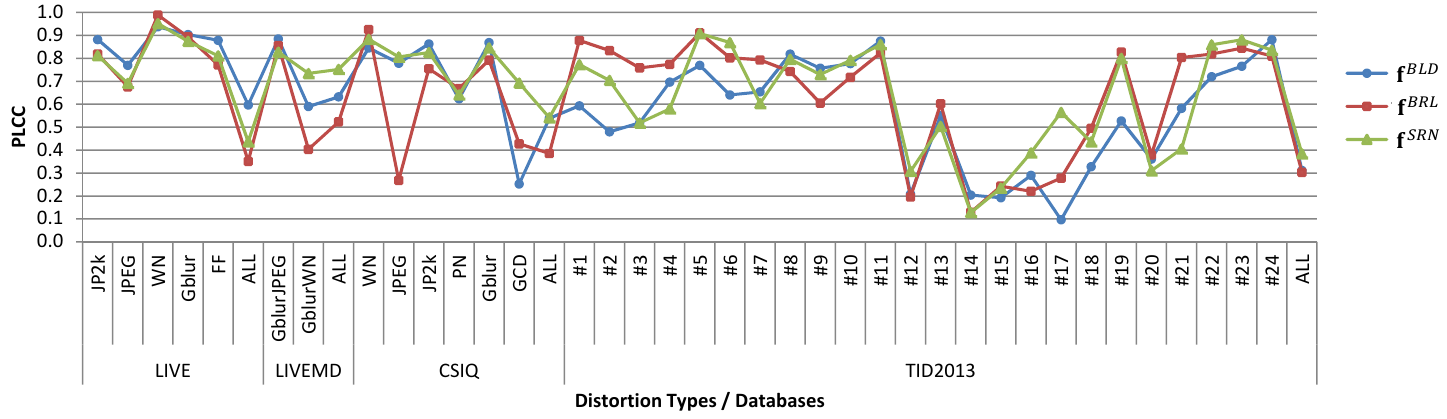}
    \vspace{-0.3cm}
    \label{feature}
    \caption{Maximum PLCCs of ${{\mathbf f}}^{BLD}\ $, ${{\mathbf f}}^{BRL}$, and ${{\mathbf f}}^{SRN}$ across all the images contained in each type of distortion and the whole image set in the LIVE, LIVEMD, CSIQ, and TID2013 databases.}
    \vspace{-0.3cm}
    \end{figure*}

\vspace{-0.3cm}
\subsection{Preparation of Training Data}
\label{sec:Preparation of Training Data}

In line with the discussions in Section \ref{sec:Generation of Preference Image Pairs}, we first collect a number of PIPs with valid preference labels. We then calculate the features of each image and the difference feature vector of each PIP. Thus the training data comprises:
\begin{enumerate}
    \item \textit{n} images: ${\mathcal I}{\rm =}\left\{I_1{\rm ,\ ...,\ }I_n\right\}$;
    \item for each image an integrated feature vector ${{\mathbf f}}^{all}_i\in {{\mathbb R}}^d{\rm ,\ }i{\rm =1,..,}n$, $d$ is the dimension of the feature vector;
    \item $N$ PIPs generated from the \textit{n }images. The PIP set is denoted as ${\mathcal P}$: ${\rm \ }{\mathcal P}\subseteq {\rm \ }\left\{\left(I_i,I_j\right)\mathrel{\left|\vphantom{\left(I_i,I_j\right) i,j=1,\dots n}\right.\kern-\nulldelimiterspace}i,j=1,\dots n\right\}$; and
    \item for each PIP $\left(I_i,I_j\right)\in {\mathcal P}$,\textit{ }a difference feature vector ${{\mathbf x}}_k$\textit{ }and a preference label $y_k,\ k=1,\dots ,N$:
    \begin{equation}
        {{\mathbf x}}_k={{\mathbf f}}^{all}_i-{{\mathbf f}}^{all}_j,
    \end{equation}
    \noindent and
    \begin{equation}
        y_k\in \left\{-1,+1\right\}.
    \end{equation}
\end{enumerate}

Considering that if an image pair $\left(I_i,I_j\right)$ corresponds to ${{\mathbf x}}_k$ and $y_k$, there is a pair $\left(I_j,I_i\right)$ associated with $-{{\mathbf x}}_k$ and $-y_k$, thus the classifier should take this symmetry into account. Let ${\mathcal X}$ be the set of difference vectors and ${\mathcal Y}$\textbf{ }the set of preference labels, such that
    \begin{equation}
        {\mathcal X}{\mathbf =}\left\{{{\mathbf x}}_1,\dots ,{{\mathbf x}}_N,{{\mathbf x}}_{N+1},\dots ,{{\mathbf x}}_{2N}\right\}\subset {{\mathbb R}}^{2N\times d},
    \end{equation}
    \noindent and
    \begin{equation}
        {\mathcal Y}{\mathbf =}\left\{y_1,\dots ,y_N,y_{N+1},\dots ,y_{2N}\right\}\subset {\left\{{\rm -}{\rm 1,+1}\right\}}^{2N},
    \end{equation}
\noindent where ${{\mathbf x}}_{N+k}=-{{\mathbf x}}_k$, $y_{N+k}=-y_k$. The mapping from difference feature vectors to preference labels can then be realized by a binary classifier trained on $\left\{{\mathcal X},{\mathcal Y}\right\}$.

\vspace{-0.3cm}
\subsection{Preference Learning via MKLGL}

Previous discussion in Part A implies that the difference feature vector includes 8 portions in accordance with ${{\mathbf f}}^{BLD}_1$, ${{\mathbf f}}^{BLD}_2$, ${{\mathbf f}}^{BLD}_3$, ${{\mathbf f}}^{BRL}_1$, ${{\mathbf f}}^{BRL}_2$, ${\mathbf m}$, ${\mathbf v}$, and ${\mathbf e}$, each of which represents the image at a particular scale or captures one particular property. We therefore introduce multiple kernel learning (MKL) to measure the similarity of different portions of features using different kernels \cite{rakotomamonjy2008simplemkl} - \cite{mklTNNLS2014}. Of the various MKL algorithms, the MKLGL approach \cite{xu2010simple} has shown its efficiency and effectiveness across a wide range of applications \cite{gonen2011multiple}. Thus we adopt MKLGL to solve the preference learning problem in the proposed BIQA framework.

The prediction function that maps the difference feature vector to the preference label can then be represented by
\begin{equation}
    f\left({\mathbf x}\right)={\sum}^{2N}_{k=1}{\left[{\alpha }_ky_k{\sum}^M_{m=1}{{\theta }_m{{\mathbf K}}_m\left({\mathbf x},{{\mathbf x}}_k\right)}\right]}.
\end{equation}
\noindent where the optimal parameters ${{\mathbf \alpha }}={\left\{{\alpha }_k\right\}}^{2N}_{k=1}$ and ${\mathbf \theta }={\left\{{\theta }_m\right\}}^M_{m=1}$ are learned from the training data $\left\{{\mathcal X},{\mathcal Y}\right\}$; ${{\mathbf K}}_m,\ m=1,\dots ,M$, is a base kernel and defines a feature mapping from the original input space to a reproducing kernel Hilbert space (RKHS) ${{\mathcal H}}_m$.

Base kernels may be diverse in the type or parameters of the kernel function, or in the portion of features on which the kernel function operates \cite{rakotomamonjy2008simplemkl}. In our research, we construct Gaussian kernels with 5 different bandwidths ($\left\{2^{-2},2^{-1},\dots ,2^2\right\}$) on each portion of the difference feature vector. In addition, we exploit Gaussian kernels with 5 different bandwidths ($\left\{2^{-2},2^{-1},\dots ,2^2\right\}$) using the entire difference feature vector to encode the potential correlations between different portions of features. In sum, we have 45 kernels (8 portions of features $\times $5 bandwidths Gaussian kernels + 5 bandwidths Gaussian kernels). We construct a kernel in this way according to \cite{xu2010simple}\cite{gonen2011multiple} to avoid using a large memory.

The learned MKLGL model can estimate the preference label of a given pair of images based on the corresponding difference feature vector ${{\mathbf x}}_{test}$. In addition, if ${{\mathbf x}}_{test}={\mathbf 0}$, we consider the two images in the test pair to be of the same quality, and assign the predicted preference label as 0.

\vspace{-0.3cm}
\subsection{Quality Prediction}
Consider a full round of comparisons of all the training images. We term the sum of the ideal preference labels ${\left\{y_{i,j}\right\}}^n_{j=1,j\ne i}\in {\left\{1,-1\right\}}^{n-1}$ associated with ${\left\{(I_i,I_j)\right\}}^n_{j=1,j\ne i{\rm \ }}$ as the ''\textit{gain}'' of $I_i$ and denote it as $g_i$, $i=1,\dots ,n$, such that
\begin{equation}
    g_i={\sum}^n_{j=1,j\ne i}{y_{i,j}}, \text{and} {-\left(n-1\right)\le g}_i\le \left(n-1\right),\ \forall i.
\end{equation}

The gain of a training image is proportional to its perceived quality, because $y_{i,j}$ essentially reflects the quality of $I_i$ relative to $I_j$, and $y_{i,j}=1$ indicates $I_i$ is better than $I_j$ in terms of quality. For simplicity, we assume a linear mapping between the gain value and the ideal perceived quality scores of the training images: $q_i=ag_i+b$, where $q_i$ is the quality score of $I_i$. Because neither all the ideal preference labels nor the ideal quality scores are available in the scenario of the proposed BIQA framework, we investigate to estimate the parameters $a$ and $b$ based on the following two assumptions:
\begin{enumerate}
\item  The training images cover the full range of possible quality, from the poorest (extremely annoying) to the best (undistorted), with small steps; and
\item  The gain value $\left(n-1\right)$ corresponds to the greatest quality score, and $-\left(n-1\right)$ corresponds to the lowest score.
\end{enumerate}

These assumptions are reasonable because it is easy to collect/generate sufficient images that diverse greatly in perceived quality, in which each pair includes an undistorted image and a heavily distorted one. Without loss of generalization, we choose the continuous quality scale of [0 100]. $a$ and $b$ can then be estimated by solving the following linear equation:
\begin{equation}
\left[ \begin{array}{cc}
n-1 & 1 \\
-(n-1) & 1 \end{array}\right]
\left[ \begin{array}{c}
a \\
b \end{array}\right]
=\left[ \begin{array}{c}
100 \\
0 \end{array}\right].
\end{equation}
\noindent We have $a={50}/{(n-1)}$ and $b=50$.

Given a test image $I_t$, we then pair it with each training image, and calculate the corresponding difference feature vector:
\begin{equation}
    {{\mathbf x}}_{t,i}={{\mathbf f}}^{all}_t-{{\mathbf f}}^{all}_i
\end{equation}
\noindent where ${{\mathbf f}}^{all}_t$ is the integrated feature vector of the test image, ${{\mathbf f}}^{all}_i$ is the feature vector of the $i$ th training image $I_i$, $,\ i=1,2,\dots ,n$. The preference labels associated with the test pairs are estimated by the trained MKLGL. Let ${\widehat{{\mathcal Y}}}_t=\left\{{\hat{y}}_{t,1},\dots ,{\hat{y}}_{t,n}\right\}\in {\left\{+1,0,-1\right\}}^n$ denote the predicted preference labels. Finally, the gain of the test image is calculated by ${\hat{g}}_t=\sum^n_{i=1}{{\hat{y}}_{t,i}}$, and the quality score is then predicted by:
\begin{equation}
    {\hat{q}}_t={\rm 50}\left[{{\hat{g}}_t}/(n-1)+1\right],
\end{equation}

\noindent where ${\hat{g}}_t$ ranges from $-n$ to $n$, and n is the number of training images. The potentially minimum quality score is $-50/(n-1)$, and the potentially maximum quality score is $(100+50/(n-1))$. Thus the proposed BIQA index is bounded to the interval $[-50/(n-1),100+50/(n-1)]$. When $n$ is fixed, the range of the predicted quality scores is fixed.

We design the BIQA index in this way in case of the absence of the potentially most/least annoying image in the training set. In most cases, the predicted quality score lies in the interval $[0,100]$. But when the test image is perceptually worse than the poorest training image, the predicted quality score becomes $-50/(n-1)$. Conversely, when the test image is better than all the training images, the predicted quality score becomes $100+50/(n-1)$, which indicates the quality improvement. This feature is similar to visual information fidelity (VIF) \cite{vif}, a full-reference image quality assessment method, whose value may become larger than 1 to indicate quality improvement.

In practical applications, we can gradually extend the training set with the emerging images whose quality scores are beyond $[0,100]$, in order to avoid the absence of the potentially most/least annoying image in the training set. In addition, we can use the logistic regression recommended by ITU \cite{bt2002500} to rescale the predicted quality scores into the interval $[0,100]$.

Even though this quality prediction approach seems rather ad-hoc at first sight, it correlates highly with human perceptions of quality, as verified by the thorough experiments presented in the following two sections. In addition,because MOS/DMOS values allow a much finer description of image quality than PIPs, and the reliability of the MOS/DMOS values in existing databases have been improved by the researchers, it would be a success if the BIQA model trained on the PIPs performs competitively with those trained on the MOS/DMOS values.

\vspace{-0.3cm}
\section{IQA Experiments}
\label{sec:IQA Experiments}
We conduct a series of experiments on the four largest datasets, i.e. the LIVE dataset, TID2013, CSIQ, and LIVEMD databases, with five objectives.

1) The first objective is to show how the performance of the proposed BIQA method varies with the parameters included in it. This is illustrated through the experiment on the LIVE database, as presented in Part A.

2) The second objective is to analyze the variability of the predicted scores across different sessions. This is verified by the experiments where we train and test the BIQA models on the LIVE database. Details are described in Part B.

3) The third objective is to evaluate the effectiveness of our BIQA model. This is verified by the experiments where we train and test the BIQA models on two distinct subsets of each database. Details are described in Part C.

4) The fourth objective is to show that the proposed BIQA approach is database independent. In this case, we train the model on the whole LIVE database and then test it on the other databases, as introduced in Part D.

5) The fifth objective is to show that it is easy to extend the proposed BIQA framework to emerging distortion categories. This is demonstrated by the experiment in which we aggregate the PIPs separately extracted from the LIVE, CSIQ, and TID2013 databases into a single training set for learning a robust model. Details are presented in Part E.

We compare the performance of the proposed BIQA method with state-of-the-art BIQA algorithms which have been reported as having the best performance \cite{mittal2012BRISQUE-L}\cite{ye2012CORNIA}, i.e., BLIINDS-II \cite{saad2012BLIINDS-II}, BRISQUE \cite{MittalBRISQUE}, BRISQUE-L \cite{mittal2012BRISQUE-L}, and CORNIA \cite{ye2012CORNIA}. In particular, we adopt the version of BLIINDS-II that utilizes support vector regression (SVR) with a radial basis function (RBF) kernel to model the relationship between ${{\mathbf f}}^{BLD}$ and quality scores. All of these algorithms are implemented based on the codes provided by the authors and the corresponding literature.

To verify the efficacy of adopting a combination of NSS features, we implement two BIQA algorithms that use ${{\mathbf f}}^{SRN}$ and ${{\mathbf f}}^{all}$ to represent an image, respectively, and utilize SVR with a RBF kernel to learn a quality prediction function. For convenience, we refer to them as ${{\mathbf f}}^{SRN}{\footnotesize \sf{SVR}}$ and ${{\mathbf f}}^{all}{\footnotesize \sf{SVR}}$, respectively. It is worth declaring that the performance of ${{\mathbf f}}^{SRN\ }{\small \sf{SVR}}$ is somewhat better than SRNSS \cite{he2012SRNSS}, thus we do not report the results of SRNSS in this paper owing to space constraints.  In addition, to verify the dependency between ${{\mathbf f}}^{all}$ and the MKLGL approach, we test the performance of 1) the proposed framework using a support vector machine (SVM) with a RBF kernel, ${\footnotesize \textsf{Proposed}}^{\footnotesize \textsf{SVM}}$,  and 2) MKLGL trained on the MOS/DMOS values using ${{\mathbf f}}^{all}$ to represent an image, ${{\mathbf f}}^{all}{\footnotesize \sf{MKL}}$.

In our experiments, the LIBrary for Support Vector Machines (LIBSVM) package \cite{LIBSVM} is used to implement the SVR and SVM algorithms. The optimal parameters of SVR and SVM algorithms are learned by 5-fold cross validation on the training set. MKLGL is implemented by the MKL toolbox \cite{gonen2011multiple} with default settings.

To evaluate the performances of the proposed method, three indexes are adopted as the criteria between the predicted quality scores by the BIQA algorithm and DMOS/MOS: \textit{Kendall's Rank Correlation Coefficient} (KRCC), \textit{Pearson's Linear Correlation Coefficient }(PLCC), and \textit{Spearman's Rank Correlation Coefficient }(SRCC). Greater KRCC, PLCC and SRCC values indicate better consistency with human opinions of quality. We perform a nonlinear mapping using a logistic function as described in \cite{video2000final} before computing these indexes.

In all the experiments, we conducted the train-test procedures 100 times to verify the robustness of the proposed method. In Parts A, B, and D, the database in each train-test procedure was randomly split into distinct training and test subsets and $N$ PIPs were randomly generated from the training images for training the proposed BIQA model. In Part C, we randomly produced $N$ PIPs from the entire LIVE database and used all the images contained in the TID2013, CSIQ, and LIVEMD databases as the test set in each train-test procedure. All the other BIQA methods adopt the training images as the instances in learning, thus we only needed to run the train-test procedure once for them in Part C. The median performance indexes across the 100 trials were reported for comparison.

\vspace{-0.3cm}
\subsection{Variation with Algorithm Parameters}
\label{sec:Variation with Algorithm Parameters}

In this subsection, we focus on the impact of choosing different algorithm parameters: 1) the threshold $T$ for generating PIPs from existing IQA databases, 2) the number of training images, $n$, and 3) the number of PIPs, $N$. We randomly choose $N_g,\ \ 1\le N_g\le 28$, groups of images contained in the LIVE database for training, and use the rest for testing. Thus $n\approx \left\lfloor {808\times N_g}/{29}\right\rfloor .$ In the experiments, we considered the following parameters: $N_g\in \left\{5,10,15,20\right\},$ $T=0,10,\dots ,70$, and $N\in \left\{500,1000,1500,2000\right\}$. It is worth noting that the number of PIPs adopted for training is $N_p={\rm min}\left(N,M\right)$, where $M$ is the maximum number of pairs we can generate in accordance with certain $T$ and $N_g$. We report the proximate $M$ in the LIVE databases in Table I. It is clear that in most cases, we only utilized a very small portion of all the possible pairs for training the proposed BIQA model.

\begin{table}[!t]
\centering
\caption{Maximum Number of Pairs That Can Be Generated ($\times {10}^3$).}
\label{NumPair}
\begin{tabular}{c|c|c|c|c|c|c|c|c}\hline \hline
$N_g$ & \multicolumn{8}{|c}{$T$} \\ \hline & 0 & 10 & 20 & 30 & 40 & 50 & 60 & 70 \\ \hline
5 & 9.3 & 7.6 & 5.9 & 4.3 & 3.0 & 2.0 & 1.2 & 0.6 \\ \hline
10 & 38.2 & 31.1 & 24.4 & 18.3 & 13.1 & 8.8 & 5.5 & 3.2 \\ \hline
15 & 86.7 & 70.4 & 55.2 & 41.4 & 29.7 & 19.9 & 12.4 & 7.1 \\ \hline
20 & 155.4 & 125.9 & 98.9 & 74.4 & 53.5 & 36.1 & 22.5 & 12.8 \\ \hline
29 & 326.0 & 264.5 & 207.6 & 156.2 & 111.9 & 75.7 & 47.1 & 27.0 \\ \hline\hline
\end{tabular}
\end{table}

\begin{figure*}[t]
\centering
\includegraphics[width=6.5in]{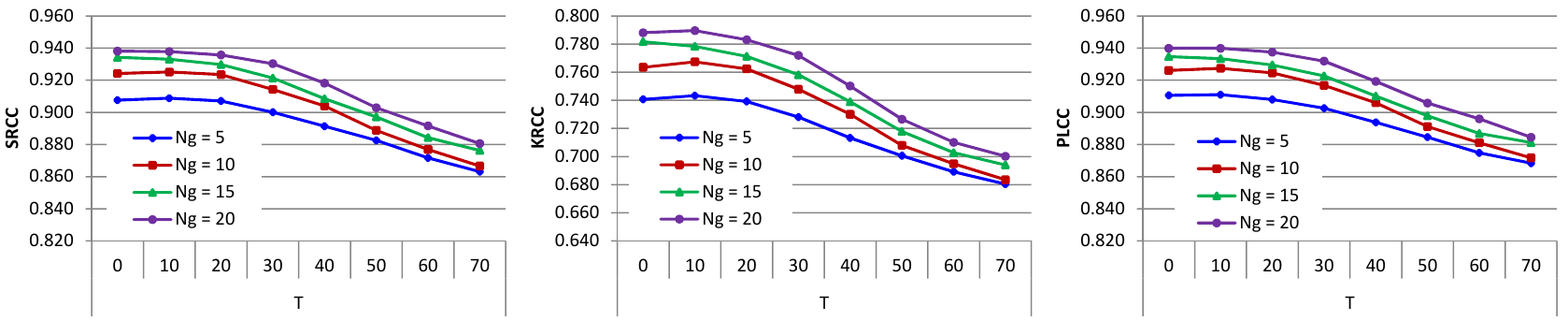}
\caption{Effect of the threshold $T$ used to generate PIPs from the LIVE database.}
\label{figure8}
\end{figure*}

\begin{figure*}[th]
\centering
\includegraphics[width=6.5in]{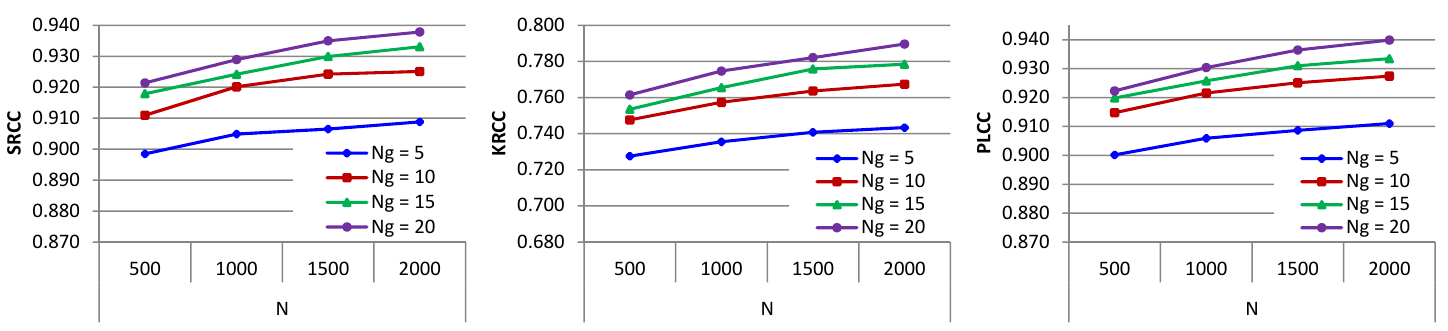}
\caption{Effect of the number of images and the number of PIPs contained in the training set (tested on the LIVE database).}
\label{figure9}
\end{figure*}

Fig. \ref{figure8} shows how the performance on the entire test set varies with the settings of $T$ when $N_g=20$ and $N=2000$. As is clear, the values of all the criteria first slightly increase and then gently descend as the thresholds become greater. This is mainly because we can reduce the noisy data contained in the training set by merely choosing the pairs whose quality scores differ sufficiently. However, when the thresholds become too great, the loss of information caused by abandoning image pairs plays a more significant role. As a result, the performance of the proposed BIQA method decreases. It is inspiring that the performance of the proposed BIQA method is acceptable even when $T=70$. Similar phenomena were observed under other settings of $N_g$ and $N$ as well as across various IQA databases.

To verify the effect of $n$ ($N_g$) and $N$, we fixed $T=10$. As is shown in Fig. \ref{figure9}, the performance of the proposed BIQA method monotonously increases as the number of images or PIPs included in the training set rises. Moreover, the number of images has more significant impact on the performance than the number of PIPs. For example, the performance when $N_g=20$ and $N=500$ is much better than when $N_g=5$ and $N=2000$. This is mainly because more comparisons are conducted for a test image when there are more training images, leading to a more precise estimation of its quality score. This result suggests that when researchers try to construct a PIP dataset, they can collect many training images but only randomly label a very small portion of all the possible image pairs.

\subsection{Variability of Scores Across Sessions}
\label{sec:Variability of Scores Across Sessions}

\begin{figure}[t]
\centering
\includegraphics[width=3in]{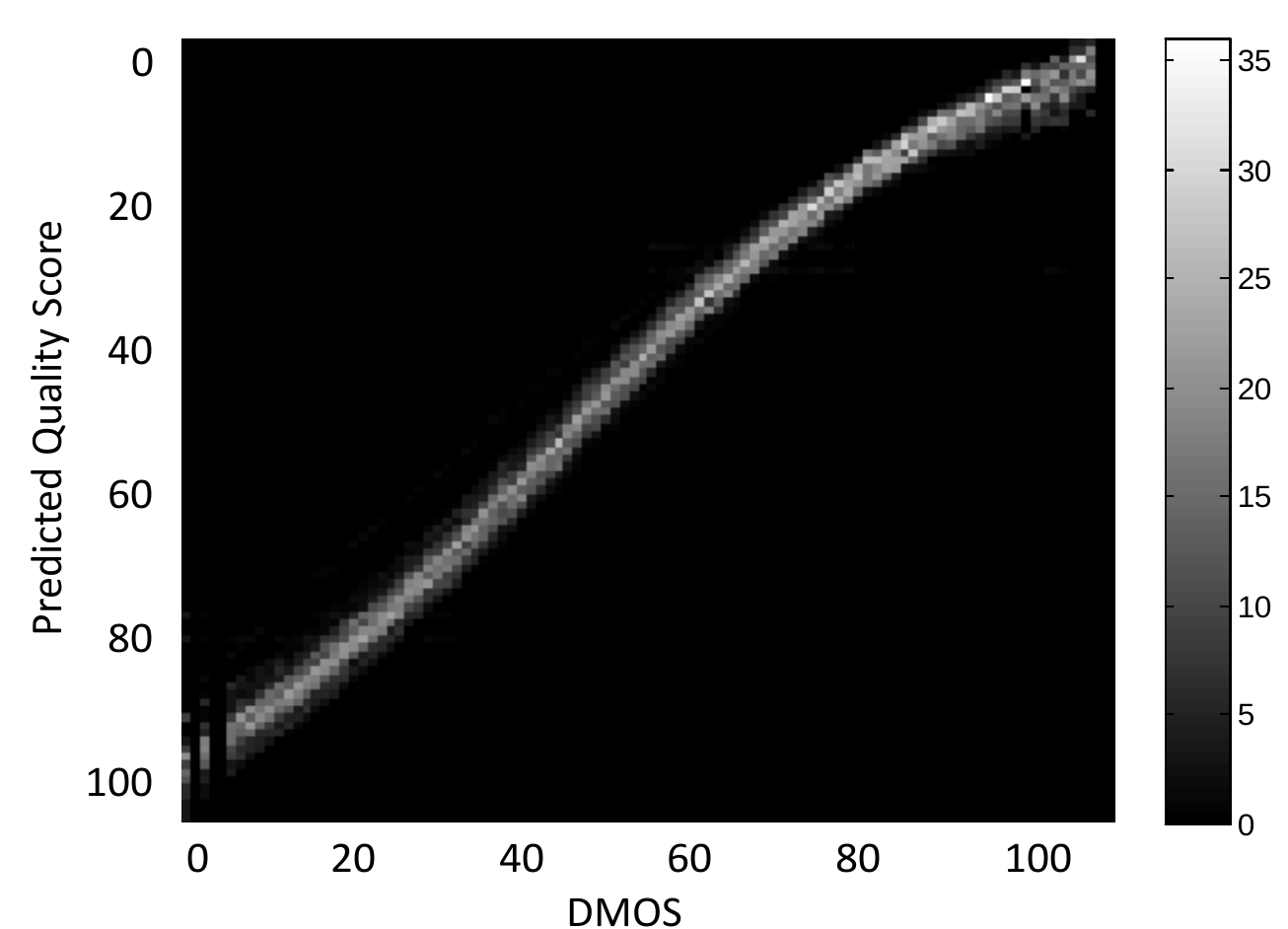}
\caption{The joint probability histogram of the predicted quality scores and the DMOS on the LIVE database across 100 sessions.}
\label{VariabilityScores}
\end{figure}

In this subsection, we focus on the variability of the predicted scores across different sessions. In particular, we estimate the joint probability histogram of the predicted quality scores and the difference mean opinion scores (DMOS) of all the test images across all the sessions. Fig. \ref{VariabilityScores} plots the joint probability histogram of the predicted quality scores and the DMOS on the LIVE database across 100 sessions, where the probability (\%) is shown in terms of pixel intensity. The experiment settings are: $N_g=20$, $N=2000$, and $T=10$. It is worth to mention that better quality is indicated by a higher predicted score, but a lower DMOS value. Thus the predicted quality score decreases as the value of DMOS increases, as shown in Fig. \ref{VariabilityScores}.

We see from Fig. \ref{VariabilityScores} that the predicted quality scores change slightly between sessions (less than 5 points in most cases). In addition, the predicted quality scores show high consistency with DMOS across all the sessions. Thus we can conclude that our method is highly effective and robust across all the sessions.

\subsection{Consistency with Human Opinions}
\label{sec:Consistency with Human Opinions}

We utilized all four databases to test the performance of the proposed BIQA method and compare it with BLIINDS-II, BRISQUE-L, CORNIA, ${{\mathbf f}}^{SRN}{\footnotesize \sf{SVR}}$, ${{\mathbf f}}^{all}{\footnotesize \sf{SVR}}$, ${{\mathbf f}}^{all}{\footnotesize \sf{MKL}}$, and ${\footnotesize \textsf{Proposed}}^{\footnotesize \textsf{SVM}}$.

On the LIVE database, we chose $N_g\in \left\{5,10,15,20\right\}$ groups of images for training and the remaining images for the test set. The threshold and number of PIPs were fixed as $T={\rm 10}$ and $N={\rm 2000}$. The performance indexes of these BIQA methods on the entire test dataset are tabulated in Table \ref{pfmLIVE}. In addition, the best BIQA method for each index is highlighted in boldface.

\begin{table}[t]
\centering
\setlength{\tabcolsep}{0.07in}
\caption{Overall Performance of BIQA Methods on the LIVE Database.}
\label{pfmLIVE}
\begin{tabular}{l|ccc|ccc} \hline \hline
                                  & \multicolumn{3}{|c|}{$N_g=5$} & \multicolumn{3}{|c}{$N_g=10$} \\
                                  & SRCC  & KRCC  & PLCC  & SRCC  & KRCC  & PLCC  \\ \hline
BRISQUE-L                         & 0.872 & 0.689 & 0.883 & 0.919 & 0.755 & 0.923 \\
BLIINDS-II                        & 0.849 & 0.663 & 0.858 & 0.885 & 0.706 & 0.892 \\
${{\mathbf f}}^{SRN}{\footnotesize \sf{SVR}}$  & 0.890 & 0.714 & 0.894 & 0.915 & 0.749 & 0.918 \\
${{\mathbf f}}^{all}{\footnotesize \sf{SVR}}$  & 0.907 & 0.741 & 0.913 & 0.927 & 0.774 & 0.932 \\
${{\mathbf f}}^{all}{\footnotesize \sf{MKL}}$  & \textbf{0.909} & 0.741 & \textbf{0.915} & \textbf{0.930} & \textbf{0.776} & \textbf{0.934} \\
CORNIA                            & 0.908 & 0.733 & 0.909 & 0.931 & 0.767 & 0.930 \\
{\footnotesize \textsf{Proposed}}        & \textbf{0.909} & \textbf{0.743} & 0.911 & 0.925 & 0.767 & 0.927 \\
${\footnotesize \textsf{Proposed}}^{\footnotesize \textsf{SVM}}$    & 0.883 & 0.706 & 0.888 & 0.914 & 0.747 & 0.917 \\ \hline \hline

                                  & \multicolumn{3}{|c|}{$N_g=15$} & \multicolumn{3}{|c}{$N_g=20$} \\
                                  & SRCC  & KRCC  & PLCC  & SRCC  & KRCC  & PLCC \\ \hline
BRISQUE-L                         & 0.934 & 0.780 & 0.938 & 0.942 & 0.795 & 0.945 \\
BLIINDS-II                        & 0.913 & 0.750 & 0.920 & 0.927 & 0.771 & 0.930 \\
${{\mathbf f}}^{SRN}{\footnotesize \sf{SVR}}$  & 0.927 & 0.772 & 0.929 & 0.938 & 0.787 & 0.940 \\
${{\mathbf f}}^{all}{\footnotesize \sf{SVR}}$  & \textbf{0.938} & \textbf{0.790} & 0.940 & \textbf{0.946} & 0.803 & \textbf{0.949} \\
${{\mathbf f}}^{all}{\footnotesize \sf{MKL}}$  & \textbf{0.938} & \textbf{0.790} & \textbf{0.941} & \textbf{0.946} & \textbf{0.807} & \textbf{0.949} \\
CORNIA                            & 0.936 & 0.777 & 0.936 & 0.942 & 0.787 & 0.942 \\
{\footnotesize \textsf{Proposed}}        & 0.933 & 0.778 & 0.933 & 0.938 & 0.790 & 0.940 \\
${\footnotesize \textsf{Proposed}}^{\footnotesize \textsf{SVM}}$    & 0.924 & 0.763 & 0.929 & 0.930 & 0.774 & 0.932 \\ \hline \hline
\end{tabular}
\end{table}

\begin{table}[t]
\centering
\setlength{\tabcolsep}{0.15in}
\caption{Settings for Constructing Training Data In Each Database.}
\label{Settings}
\begin{tabular}{l|c|c|c|c} \hline \hline
           & $N_g$ & $N$  & MOS/DMOS Scale & $T$ \\ \hline
LIVE       & 20    & 2000 & (-3, 112)      & 10  \\
CSIQ       & 20    & 2000 & [0, 1]         & 0.1 \\
TID2013    & 17    & 2000 & (0.2, 7.3)     & 1   \\
LIVEMD     & 10    & 2000 & (0, 85)        & 10  \\ \hline \hline
\end{tabular}
\end{table}

\newcounter{MYtemptabcnt}
\begin{table*}[!t]
\centering
\footnotesize
\setcounter{MYtemptabcnt}{\value{table}}
\setcounter{table}{3}
\newcommand{\dbcol}{\multicolumn{1}{c|}}
\newcommand{\dbcolI}{\multicolumn{2}{c}}
\newcommand{\colc}{\multicolumn{1}{c}}
\caption{Performance of BIQA Methods Across Various Distortion Categories of The TID2013, LIVE, CSIQ, LIVEMD Databases.}
\label{pfmALL}
\setlength{\tabcolsep}{0.06in}
\begin{tabular}{l|cccccccccc@{\hspace{-0.01in}}c@{\hspace{-0.01in}}ccccc@{\hspace{-0.01in}}c} \hline \hline
 & \multicolumn{17}{c}{TID2013} \\ 
 & \#1 & \#2 & \#3 & \#4 & \#5 & \#6 & \#7 & \#8 & \#9 & \#10 & \#11 & \#19 & \#21 & {\#22} & \#23 & \#24 & ALLsub \\ \hline
BRISQUE-L & {\textbf{0.808}} & \textbf{0.682} & \textbf{0.815} & \textbf{0.489} & \textbf{0.871} & \textbf{0.847} & 0.744 & 0.853 & 0.652 & 0.831 & 0.818 & \textbf{0.787} & 0.652 & {0.860} & 0.757 & 0.864 & \colc{0.759} \\
BLIINDS-II & 0.543 & 0.444 & 0.602 & 0.335 & 0.700 & 0.632 & 0.641 & 0.825 & 0.744 & 0.806 & 0.854 & 0.515 & 0.174 & {0.733} & 0.614 & 0.830 & \colc{0.615} \\
${{\mathbf f}}^{SRN}{\footnotesize \sf{SVR}}$  & 0.686 & 0.534 & 0.637 & 0.459 & 0.846 & 0.820 & \textbf{0.790} & \textbf{0.922} & 0.796 & \textbf{0.870} & 0.895 & 0.685 & 0.472 & {0.853} & \textbf{0.786} & \textbf{0.920} & \colc{0.750} \\
${{\mathbf f}}^{all}{\footnotesize \sf{SVR}}$ & 0.744 & 0.573 & 0.765 & 0.392 & 0.851 & 0.787 & 0.753 & 0.878 & 0.767 & 0.841 & 0.887 & 0.703 & 0.677 & {0.859} & 0.726 & 0.898 & \colc{0.771} \\
${{\mathbf f}}^{all}{\footnotesize \sf{MKL}}$ & 0.788 & 0.583 & 0.784 & 0.377 & 0.882 & 0.822 & 0.771 & 0.875 & 0.766 & 0.850 & 0.902 & 0.750 & 0.577 & {\textbf{0.867}} & 0.731 & 0.911 & \colc{0.764} \\
CORNIA & 0.666 & 0.511 & 0.772 & 0.424 & 0.789 & 0.624 & 0.747 & 0.876 & \textbf{0.808} & 0.841 & 0.824 & 0.617 & \textbf{0.775} & {0.766} & 0.719 & 0.892 & \colc{\textbf{0.779}} \\
{\footnotesize \textsf{Proposed}} & 0.701 & 0.564 & 0.692 & 0.409 & 0.838 & 0.776 & 0.665 & 0.898 & 0.776 & 0.832 & 0.901 & 0.620 & 0.615 & {0.815} & 0.754 & 0.892 & \colc{\textbf{0.779}} \\
${\footnotesize \textsf{Proposed}}^{\footnotesize \textsf{SVM}}$ & 0.716 & 0.584 & 0.685 & 0.379 & 0.844 & 0.840 & 0.717 & 0.894 & 0.754 & 0.873 & \textbf{0.912}  & 0.637 & 0.578 & {0.833} & 0.767 & 0.889 & \colc{0.765} \\ \hline \hline

 & \multicolumn{6}{c}{LIVE} & \multicolumn{5}{|c}{CSIQ} & \multicolumn{5}{|c|}{LIVEMD} & \multirow{2}{*}{\centering Avg.} \\ 
 & JP2k & JPEG & WN & Gblur & FF & \dbcol{ALL} & WN & JPEG & JP2k & Gblur & \dbcol{ALLsub}  & \dbcolI{GblurJPEG} & \dbcolI{GblurWN} & \dbcol{ALL} \\ \hline
BRISQUE-L & 0.924 & 0.951 & 0.979 & 0.952 & 0.900 & \dbcol{0.942} & 0.382 & 0.642 & 0.658 & 0.688 & \dbcol{0.602} & \dbcolI{0.889} & \dbcolI{0.860} & \dbcol{0.869}  & 0.786 \\
BLIINDS-II & 0.931 & 0.941 & 0.947 & 0.934 & 0.886 & \dbcol{0.927}  & 0.868 & 0.890 & 0.850 & 0.860 & \dbcol{0.864} & \dbcolI{0.815} & \dbcolI{0.827} & \dbcol{0.812}  & 0.742 \\
${{\mathbf f}}^{SRN}{\footnotesize \sf{SVR}}$  & 0.920 & 0.943 & 0.964 & 0.952 & 0.863 & \dbcol{0.938}  & \textbf{0.885} & \textbf{0.903} & \textbf{0.893}  & \textbf{0.890} & \dbcol{\textbf{0.896}} & \dbcolI{0.860} & \dbcolI{0.881} & \dbcol{0.873}  & 0.826 \\
${{\mathbf f}}^{all}{\footnotesize \sf{SVR}}$ & 0.933 & \textbf{0.956} & 0.980 & \textbf{0.955} & \textbf{0.914} & \dbcol{\textbf{0.946}} & 0.832 & 0.847 & 0.814 & 0.795 & \dbcol{0.813}  & \dbcolI{0.879} & \dbcolI{\textbf{0.902}} & \dbcol{0.889}  & 0.828 \\
${{\mathbf f}}^{all}{\footnotesize \sf{MKL}}$ & 0.933 & 0.955 & \textbf{0.981} & 0.952 & 0.891 & \dbcol{\textbf{0.946}} & 0.890 & 0.865 & 0.861 & 0.842  & \dbcol{0.855} & \dbcolI{0.879} & \dbcolI{0.900} & \dbcol{0.889}  & 0.831 \\
CORNIA & 0.920 & 0.938 & 0.963 & 0.954 & \textbf{0.914} & \dbcol{0.942}  & 0.670 & 0.859 & 0.859 & 0.813 & \dbcol{0.827}  & \dbcolI{\textbf{0.895}} & \dbcolI{0.896} & \dbcol{\textbf{0.895}}  & 0.834 \\
{\footnotesize \textsf{Proposed}} & \textbf{0.944} & 0.945 & 0.973 & 0.953 & 0.908 & \dbcol{0.938}  & 0.806 & 0.842 & 0.858 & 0.838 & \dbcol{0.843}  & \dbcolI{0.894} & \dbcolI{0.898} & \dbcol{\textbf{0.895}}  & \textbf{0.836} \\
${\footnotesize \textsf{Proposed}}^{\footnotesize \textsf{SVM}}$ & 0.917 & 0.935 & 0.970 & 0.932 & 0.876 & \dbcol{0.930}  & 0.746 & 0.831 & 0.845 & 0.826 & \dbcol{0.824}  & \dbcolI{0.895} & \dbcolI{0.900} & \dbcol{0.887}  & 0.824 \\ \hline \hline
\end{tabular}
\setcounter{table}{\value{MYtemptabcnt}}
\end{table*}

It is notable that the performance of the proposed method is much better than BRISQUE-L, BLIINDS-II, and ${{\mathbf f}}^{SRN}{\footnotesize \sf{SVR}}$ when $N_g=5$. When the number of training images increases, it still performs better than BLIINDS-II, and ${{\mathbf f}}^{SRN}{\footnotesize \sf{SVR}}$ and is highly comparable to the other algorithms. Given that the proposed method utilizes much less information than the other methods which use DMOS in learning (as little as $1.3\%$ of all the possible image pairs when $T={\rm 10}$ and $N_g={\rm 20}$), this is a solid achievement.

The fact that both ${{\mathbf f}}^{all}{\small \sf{SVR}}$ and ${{\mathbf f}}^{all}{\small \sf{MKL}}$ perform consistently better than BRISQUE-L, BLIINDS-II, and ${{\mathbf f}}^{SRN}{\small \sf{SVR}}$ verifies the effectiveness of combining existing NSS features to represent an image. Further, the algorithm using the proposed framework and MKLGL ({\small \textsf{Proposed}}) consistently outperforms that using SVM (${\small \textsf{Proposed}}^{\footnotesize \textsf{SVM}}$), demonstrating the important correlation between MKLGL and the integrated features. In addition, the performance of ${\small \textsf{Proposed}}^{\footnotesize \textsf{SVM}}$ is also comparable with state-of-the-art algorithms, corroborating the validity of the proposed BIQA framework.

For each of the CSIQ, TID2013, and LIVEMD databases, we randomly select 2/3 of all the dataset content for training and the remaining 1/3 for test in each trial. There are 6 and 24 distortion types in the CSIQ and TID2013 databases, respectively. However, the features adopted in the proposed BIQA method are not representative of all of them, as shown in Fig. 7. Thus we test all the BIQA methods only on the distortion subsets with which ${{\mathbf f}}^{all}$ correlates highly, i.e. WN, JPEG, JP2k, and Blur in the CSIQ database, and \#\# 1-11, 19, 21-24 in TID2013. The parameters for constructing the training data were chosen according to the experiments on the LIVE database as well as the quality scale in the corresponding database. Details are listed in Table \ref{Settings}.

Table \ref{pfmLIVE} shows that the SRCC, KRCC and PLCC produce unanimous comparison results, thus we only report the SRCC values in this subsection for clarity. The SRCCs of the BIQA methods for each distortion category and the entire dataset ({\small \sf{ALL}} or {\small \sf{ALLsub}}, where {\small \sf{ALLsub}} refers to "all the subsets") are listed in Table \ref{pfmALL}. The performance on the LIVE database corresponds to $N_g=20$. In addition, we compute the average SRCC over the four databases for each method and report it in the last column of the table. The average SRCC is calculated by weighting the SRCC values according to the number of images contained in each database. The best BIQA algorithm for each dataset is highlighted in boldface.

It is rather inspiring that our approach statistically yields the best result across all databases. The proposed method coupled with CORNIA shows the best performance on the TID2013 and LIVEMD databases. Moreover, across all the distortion categories of all the databases, the performance of the proposed method is highly comparable with state-of-the-art BIQA algorithms. It is notable that BRISQUE-L performs poorly on the CSIQ database. This is mainly due to the weak correlation between ${{\mathbf f}}^{BRL}$ and the quality scores of the CSIQ database, as illustrated in Fig. 7. In contrast, the proposed method combats this drawback, again demonstrating the efficacy of employing the fused NSS features.

\begin{figure*}[htp]
\centering
\includegraphics[width=7in]{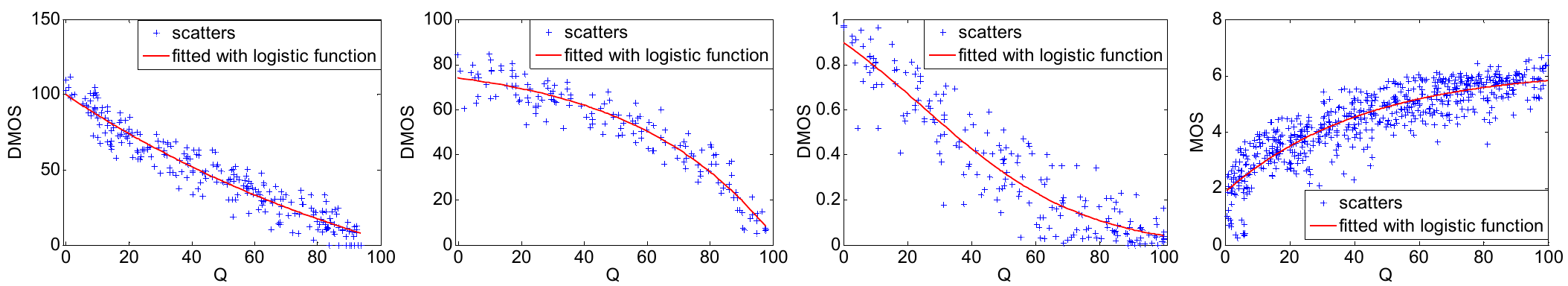}
\caption{Scatter plots of the predicted results $Q$ vs. $DMOS/MOS$ for each of the entire LIVE, LIVEMD, CSIQ, and TID2013 databases (from left to right).}
\label{scatter}
\end{figure*}

Scatter plots (for each of the entire LIVE, LIVEMD, CSIQ, and TID2013 databases) of the predicted quality scores versus DMOS/MOS on the test sets are shown in Fig. \ref{scatter}. We performed a logistic regression for each database separately, because the scales of the DMOS/MOS values in the LIVE, LIVEMD, TID2013, and CSIQ databases are different from each other. The tight clustering property and the monotonic relationship compared to DMOS/MOS demonstrate the impressive consistency between the predicted quality scores by the proposed BIQA approach and the human opinions of quality.

We report the standard deviations of SRCC values on the entire dataset over the 100 trials and the weighted averages across the four databases in Table \ref{StdPfm}. The randomness of the train-test split coupled with that of constructing PIPs leads to higher fluctuations in the performance of the proposed approach compared to the best BIQA methods. However, the standard deviation of the proposed method is still comparable with those of BLIINDS-II and ${{\mathbf f}}^{all}{\footnotesize \sf{SVR}}$. From Table \ref{pfmALL}, Fig. \ref{scatter}, and Table \ref{StdPfm}, we conclude that our method correlates highly with human perception of image quality and is insensitive to datasets.

\subsection{Database Independence}
\label{Database Independence}

Because the proposed BIQA method is learning-based, it is necessary to verify whether its performance is bound to the database on which it is trained. To show this, we trained the BIQA method on the entire LIVE database and then tested it on the TID2013, CSIQ, and LIVEMD databases. In Table \ref{PfmLIVEOthers}, we report the performance of the BIQA methods across all the distortion categories and the entire database ({\small \sf{ALL}}). In addition, the overall performance on the distortion subsets of which ${{\mathbf f}}^{all}$ is representative was evaluated ({\small \sf{ALLsub}}). Similarly, we computed the weighted average SROCCs across the three databases, and these are listed in the last two columns. In particular, the calculation of {\small \sf{Avg. ALLsub}} considered \#\# 1-11, 19, 21-24 in the TID2013, WN, JPEG, JP2k, and Blur in the CSIQ database, and GblurJPEG in the LIVEMD database.

\begin{table}[t]
\centering
\setlength{\tabcolsep}{0.1in}
\setcounter{MYtemptabcnt}{\value{table}}
\setcounter{table}{4}
\caption{Standard Deviations of The SRCC Values for The Entire Database Over 100 Trials.}
\label{StdPfm}
\begin{tabular}{l|ccccc} \hline \hline
                                    & LIVE & CSIQ & LIVEMD & TID2013 & Avg. \\ \hline
BRISQUE-L                           & \textbf{0.012} & 0.076 & 0.041 & 0.053 & 0.046 \\
BLIINDS-II                          & 0.022 & 0.030 & 0.060 & 0.084 & 0.060 \\
${{\mathbf f}}^{SRN}{\footnotesize \sf{SVR}}$    & 0.019 & 0.031 & 0.064 & \textbf{0.049} & \textbf{0.041} \\
${{\mathbf f}}^{all}{\footnotesize \sf{SVR}}$    & 0.014 & 0.041 & 0.043 & 0.064 & 0.048 \\
${{\mathbf f}}^{all}{\footnotesize \sf{MKL}}$    & 0.015 & 0.036 & 0.039 & 0.061 & 0.045 \\
CORNIA                              & \textbf{0.012} & \textbf{0.026} & \textbf{0.027} & 0.063 & 0.042 \\
{\footnotesize \textsf{Proposed}}   & 0.029 & 0.045 & 0.046 & 0.072 & 0.056 \\
${\footnotesize \textsf{Proposed}}^{\footnotesize \textsf{SVM}}$   & 0.039 & 0.048 & 0.042 & 0.070 & 0.057 \\ \hline \hline
\end{tabular}
\setcounter{table}{\value{MYtemptabcnt}}
\end{table}

\begin{table*}[!t]
\footnotesize
\setcounter{MYtemptabcnt}{\value{table}}
\setcounter{table}{5}
\centering
\caption{Performance of BIQA Methods When Trained on The LIVE database But Test on The TID2013, CSIQ, and LIVEMD Databases.}
\label{PfmLIVEOthers}
\setlength{\tabcolsep}{0.03in}
\newcommand{\dbcol}{\multicolumn{1}{c|}}
\begin{tabular}{@{\hspace{-0.003in}}l@{\hspace{-0.003in}}|cccc@{\hspace{-0.003in}}c@{\hspace{-0.003in}}cccccccc@{\hspace{-0.003in}}cccc@{\hspace{-0.01in}}c@{\hspace{-0.01in}}c@{\hspace{-0.005in}}cc}
\hline\hline
 & \multicolumn{21}{c}{TID2013} \\
 & \#1 & \#2 & \#3 & \#4 & \#5 & \#6 & \#7 & \#8 & \#9 & \#10 & \#11 & \#12 & \#13 & \#14 & \#15 & \multicolumn{2}{c}{\#16} & \#17 & \#18 & \#19 & \#20 \\ \hline
BRISQUE-L & \textbf{0.889} & \textbf{0.784} & 0.380 & 0.674 & \textbf{0.888} & 0.686 & 0.757 & 0.769 & 0.559 & 0.842 & 0.855 & 0.045 & 0.390 & 0.183 & 0.215 & \multicolumn{2}{c}{0.097} & 0.189 & 0.183 & \textbf{0.781} & 0.194 \\
BLIINDS-II & 0.788 & 0.562 & 0.369 & 0.668 & 0.802 & 0.632 & 0.518 & 0.828 & 0.703 & 0.645 & 0.721 & 0.112 & 0.305 & 0.119 & 0.251 & \multicolumn{2}{c}{0.083} & 0.052 & 0.296 & 0.731 & 0.095 \\
${{\mathbf f}}^{SRN}{\footnotesize \sf{SVR}}$  & 0.468 & 0.303 & 0.244 & 0.546 & 0.694 & 0.465 & 0.401 & 0.613 & 0.560 & 0.383 & 0.848 & 0.169 & 0.190 & 0.131 & 0.054 & \multicolumn{2}{c}{0.131} & 0.393 & 0.162 & 0.406 & 0.158 \\
${{\mathbf f}}^{all}{\footnotesize \sf{SVR}}$ & 0.511 & 0.293 & 0.257 & 0.517 & 0.694 & 0.281 & 0.496 & 0.675 & 0.602 & 0.705 & 0.753 & 0.030 & 0.027 & 0.050 & 0.170 & \multicolumn{2}{c}{\textbf{0.397}} & 0.427 & 0.271 & 0.344 & 0.225 \\
${{\mathbf f}}^{all}{\footnotesize \sf{MKL}}$ & 0.786 & 0.504 & 0.259 & 0.723 & 0.862 & 0.603 & 0.680 & 0.852 & 0.714 & 0.860 & 0.919 & 0.012 & 0.300 & 0.006 & 0.003 & \multicolumn{2}{c}{0.200} & \textbf{0.514} & 0.020 & 0.638 & 0.037 \\
CORNIA & 0.761 & 0.679 & \textbf{0.615} & \textbf{0.686} & 0.828 & 0.741 & 0.399 & \textbf{0.915} & \textbf{0.834} & \textbf{0.885} & 0.899 & \textbf{0.622} & \textbf{0.655} & \textbf{0.371} & 0.168 & \multicolumn{2}{c}{0.123} & 0.173 & 0.071 & 0.659 & \textbf{0.483} \\
{\footnotesize \textsf{Proposed}} & 0.764 & 0.727 & 0.505 & 0.664 & 0.736 & 0.732 & \textbf{0.768} & 0.818 & 0.742 & 0.873 & 0.908 & 0.105 & 0.408 & 0.082 & \textbf{0.358} & \multicolumn{2}{c}{0.208} & 0.099 & 0.332 & 0.657 & 0.096 \\
 ${\footnotesize \textsf{Proposed}}^{\footnotesize \textsf{SVM}}$ & 0.792 & 0.764 & 0.561 & \textbf{0.686} & 0.760 & \textbf{0.748} & 0.406 & 0.906 & 0.760 & 0.824 & \textbf{0.920} & 0.216 & 0.375 & 0.037 & 0.223 & \multicolumn{2}{c}{0.163} & 0.202 & \textbf{0.338} & 0.652 & 0.145 \\ \hline  \hline
 & \multicolumn{6}{c|}{TID2013} & \multicolumn{8}{c|}{CSIQ} & \multicolumn{5}{c|}{LIVEMD} & \multicolumn{2}{c}{Avg.} \\
 & \#21 & \#22 & \#23 & \#24 & ALLsub & \dbcol{ALL} & WN & JPEG & JP2k & PN & Gblur & GCD & ALLsub & \dbcol{ALL} & \multicolumn{2}{c}{GblurJPEG} & \multicolumn{2}{c}{GblurWN} & \dbcol{ALL} & ALLsub & ALL \\ \hline
BRISQUE-L & 0.738 & 0.787 & 0.693 & 0.892 & 0.565 & \dbcol{0.371} & 0.816 & 0.374 & 0.684 & 0.145 & 0.729 & 0.146 & 0.431 & \dbcol{0.314} & \multicolumn{2}{c}{0.765} & \multicolumn{2}{c}{0.239} & \dbcol{0.454} & 0.552 & 0.368 \\
BLIINDS-II & 0.574 & 0.616 & 0.670 & 0.829 & 0.602 & \dbcol{0.395} & \textbf{0.868} & 0.893 & 0.833 & 0.404 & 0.842 & 0.171 & 0.861 & \dbcol{0.565} & \multicolumn{2}{c}{0.666} & \multicolumn{2}{c}{0.547} & \dbcol{0.570} & 0.662 & 0.447 \\
${{\mathbf f}}^{SRN}{\footnotesize \sf{SVR}}$  & 0.413 & 0.538 & 0.545 & 0.808 & 0.421 & \dbcol{0.287} & 0.854 & \textbf{0.894} & 0.876 & \textbf{0.568} & 0.876 & 0.393 & \textbf{0.883} & \dbcol{\textbf{0.673}} & \multicolumn{2}{c}{0.701} & \multicolumn{2}{c}{0.603} & \dbcol{0.651} & 0.542 & 0.402 \\
${{\mathbf f}}^{all}{\footnotesize \sf{SVR}}$ & 0.562 & 0.607 & 0.573 & 0.750 & 0.445 & \dbcol{0.310} & 0.545 & 0.174 & 0.467 & 0.432 & 0.497 & \textbf{0.553} & 0.327 & \dbcol{0.368} & \multicolumn{2}{c}{0.762} & \multicolumn{2}{c}{0.637} & \dbcol{0.680} & 0.445 & 0.360 \\
${{\mathbf f}}^{all}{\footnotesize \sf{MKL}}$ & 0.566 & 0.770 & 0.744 & 0.878 & 0.542 & \dbcol{0.369} & 0.838 & 0.585 & 0.652 & 0.413 & 0.498 & 0.071 & 0.531 & \dbcol{0.467} & \multicolumn{2}{c}{0.845} & \multicolumn{2}{c}{0.602} & \dbcol{0.668} & 0.564 & 0.420 \\
CORNIA & \textbf{0.874} & 0.530 & 0.749 & 0.714 & 0.416 & \dbcol{0.289} & 0.749 & 0.891 & \textbf{0.901} & 0.414 & \textbf{0.884} & 0.302 & 0.881 & \dbcol{0.659} & \multicolumn{2}{c}{\textbf{0.858}} & \multicolumn{2}{c}{\textbf{0.847}} & \dbcol{\textbf{0.835}} & 0.550 & 0.420 \\
{\footnotesize \textsf{Proposed}} & 0.636 & \textbf{0.840} & 0.636 & 0.895 & \textbf{0.604} & \dbcol{\textbf{0.481}} & 0.824 & 0.857 & 0.885 & 0.238 & 0.845 & 0.104 & 0.848 & \dbcol{0.545} & \multicolumn{2}{c}{0.820} & \multicolumn{2}{c}{0.396\textbf{}} & \dbcol{0.515} & \textbf{0.673} & \textbf{0.497} \\
${\footnotesize \textsf{Proposed}}^{\footnotesize \textsf{SVM}}$  & 0.604 & 0.816 & \textbf{0.766} & \textbf{0.917} & 0.600 & \dbcol{0.404} & 0.828 & 0.793 & 0.624 & 0.373 & 0.666 & 0.196 & 0.571 & \dbcol{0.469} & \multicolumn{2}{c}{0.823} & \multicolumn{2}{c}{0.629} & \dbcol{0.733} & 0.612 & 0.452 \\ \hline\hline
\end{tabular}
\setcounter{table}{\value{MYtemptabcnt}}
\end{table*}

It is notable that the proposed BIQA method obtains the best performance on the TID2013, both on the subsets and the entire database. On the CSIQ database, its performance is acceptable and is comparable to CORNIA, ${{\mathbf f}}^{all}{\small \sf{SVR}}$, and ${{\mathbf f}}^{all}{\small \sf{MKL}}$. On the LIVEMD database, all the BIQA methods that employ NSS features fail to precisely predict the quality scores for the images corrupted by GblurWN. This is possibly because Gblur and WN have contrary effects on the NSS features, causing the numerical relation between the features and the quality score of GblurWN images to differ significantly from those of the singly distorted images \cite{jayaraman2012objective}. In contrast, Gblur and JPEG cause similar deviation trends on the features; thus the BIQA model learned from singly distorted images works well for GblurJPEG, and the proposed method outperforms most of the existing BIQA algorithms for GblurJPEG.

Statistically, the proposed method obtains the best SRCC values over all databases. In addition, the high consistency between the predicted quality scores and DMOS across various distortion categories and databases corroborate the expectation that the parameters are not over-fitted and the learned model is insensitive to different databases.

From the experimental results shown in Parts B-D, we can draw the following two conclusions:

1) When the model is trained and tested on the same database, ${\footnotesize \textsf{Proposed}}^{\footnotesize \textsf{SVM}}$ performs slightly worse than ${{\mathbf f}}^{all}{\footnotesize \sf{SVR}}$; {\footnotesize \textsf{Proposed}} competes favorably with ${{\mathbf f}}^{all}{\footnotesize \sf{MKL}}$. This corroborates our expectation that the robust BIQA model trained on PIPs can perform competitively with those trained on the MOS/DMOS values; and

2) When the model is trained on the LIVE database but tested on the other databases, both {\footnotesize \textsf{Proposed}} and ${\footnotesize \textsf{Proposed}}^{\footnotesize \textsf{SVM}}$ statistically outperform ${{\mathbf f}}^{all}{\footnotesize \sf{SVR}}$ and ${{\mathbf f}}^{all}{\footnotesize \sf{MKL}}$. This demonstrates that the BIQA models trained on PIPs are less dependent on the training dataset than the models trained on MOS/DMOS values.

\subsection{Easy Extension of the Proposed Method}
\label{sec:Easy Extension of the Proposed Method}

Having evaluated the effectiveness and database-independent property of our method, we now demonstrate that the proposed BIQA framework can be easily extended to emerging distortion categories by simply adding a small number of corresponding PIPs into the training set. To show this, we conducted the following two experiments.

\paragraph{On the LIVE database}
\label{sec:On the LIVE database}

\begin{figure}[t]
\centering
\includegraphics[width=3in]{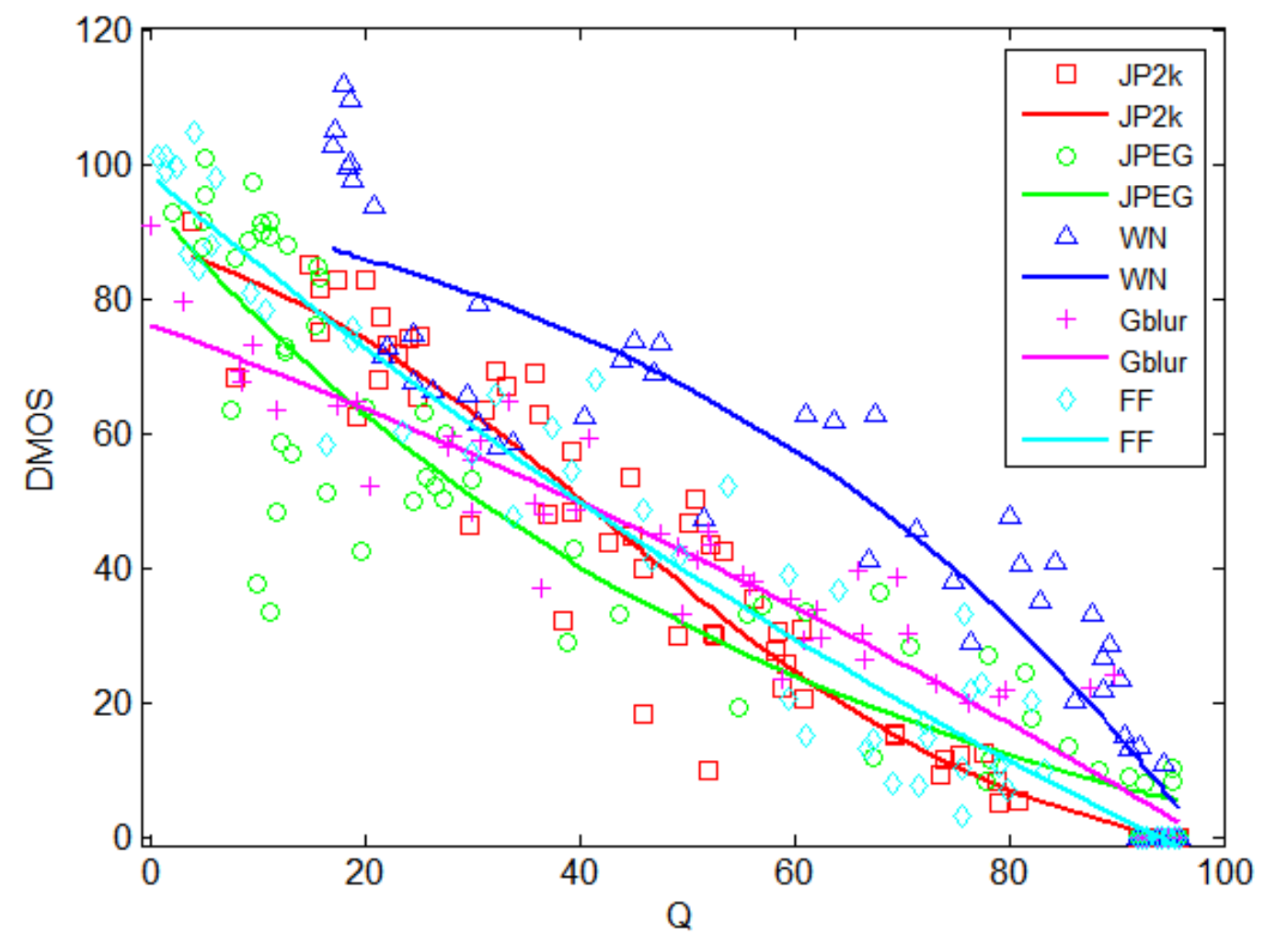}
\caption{Scatter plots of the predicted results $Q$ vs. $DMOS$ for each distortion subset in the LIVE database.}
\label{scatter_EachDis}
\end{figure}

We separately generated 400 PIPs from each distortion subset in the LIVE database with $N_g=20$ and $T=10$, and then aggregated them into a single training set in learning. Subsequently, we applied the learned BIQA model to the test images. Similar to previous experiments, the train-test procedure was repeated 100 times and the median SRCC values for each distortion category and the entire database are calculated and reported in Table \ref{ExtensionLIVE}. Scatter plots of the predicted quality scores and DMOS for each distortion category are shown in Fig. \ref{scatter_EachDis}.

From Table \ref{ExtensionLIVE} and Fig. \ref{scatter_EachDis} we conclude that our method has an impressive consistency with human perception across all the distortion subsets. The acceptable performance on the whole demonstrates that we can extend the proposed framework to various distortion types by simply adding the corresponding PIPs into the training set. It is encouraging that only hundreds of PIPs for each distortion category can lead to such a satisfactory performance. This will facilitate the application of BIQA methods since the generation of PIPs is very easy and efficient.

The overall performance is inferior to the performance when we used 2000 PIPs generated from the entire LIVE database without distinguishing the distortion categories, as shown in Table \ref{ExtensionLIVE}. This is mainly because the information about the relative quality between images afflicted by different types of distortion in the training set of this experiment is not considered, leading to quality scale mismatch errors between different distortion subsets, i.e. the predicted scores for WN diverge from the other distortion categories, as shown in Fig. \ref{scatter_EachDis}. This result suggests that, to extend the proposed BIQA method to new distortion types, it is necessary to add both intra-distortion and inter-distortion PIPs.

\begin{table}[t]
\footnotesize
\setcounter{MYtemptabcnt}{\value{table}}
\setcounter{table}{6}
\centering
\caption{Performance of The Proposed BIQA Method When PIPs Were Separately Generated from Each Distortion Subset in The LIVE Database.}
\label{ExtensionLIVE}
\setlength{\tabcolsep}{0.1in}
\begin{tabular}{l|cccccc} \hline \hline
         & JP2k  & JPEG  & WN    & Gblur & FF    & ALL   \\ \hline
{\footnotesize \textsf{Proposed}} & 0.945 & 0.936 & 0.957 & 0.949 & 0.921 & 0.875 \\ \hline \hline
\end{tabular}
\setcounter{table}{\value{MYtemptabcnt}}
\end{table}

\paragraph{On Hybrid Databases}

We also conducted experiments on the LIVE, TID2013, and CSIQ databases, which differ greatly in image content, distortion categories and quality scales. In each train-test procedure, we constructed 1000 PIPs from each database and then aggregated them into a single training set with no post-processing. The parameters $N_g$ and $T$ used to generate PIPs from each database are the same as shown in Table \ref{Settings}. Similar to previous experiments, we only considered the distortion categories in which the features were representative.

\begin{table*}[!t]
\footnotesize
\setcounter{MYtemptabcnt}{\value{table}}
\setcounter{table}{7}
\centering
\caption{Performance of BIQA Methods When The Training Data Were Collected From Hybrid Databases}
\label{ExtensionHybrid}
\newcommand{\dbcol}{\multicolumn{1}{c|}}
\setlength{\tabcolsep}{0.08in}
\begin{tabular}{l|ccccccccccccccc} \hline \hline
 & \multicolumn{15}{c}{TID2013} \\
 & \#1 & \#2 & \#3 & \#4 & \#5 & \#6 & \#7 & \#8 & \#9 & \#10 & \#11 & \#19 & \#21 & \#22 & \#23 \\ \hline
BRISQUE-L & 0.737 & 0.604 & 0.804 & 0.429 & 0.770 & 0.719 & 0.742 & 0.815 & 0.702 & 0.845 & 0.870 & \textbf{0.720} & \textbf{0.822} & 0.797 & 0.734 \\
BLIINDS-II & 0.544 & 0.320 & 0.607 & 0.425 & 0.655 & 0.590 & 0.528 & 0.780 & 0.666 & 0.779 & 0.804 & 0.536 & 0.594 & 0.701 & 0.539 \\
${{\mathbf f}}^{SRN}{\footnotesize \sf{SVR}}$  & 0.558 & 0.423 & 0.678 & 0.309 & 0.763 & 0.686 & 0.708 & 0.875 & 0.786 & 0.776 & 0.847 & 0.570 & 0.448 & 0.777 & \textbf{0.770} \\
${{\mathbf f}}^{all}{\footnotesize \sf{SVR}}$ & 0.612 & 0.449 & 0.718 & 0.326 & 0.750 & 0.697 & 0.689 & 0.865 & 0.739 & 0.818 & 0.871 & 0.577 & 0.730 & 0.806 & 0.721 \\
CORNIA & 0.726 & 0.582 & \textbf{0.720} & \textbf{0.553} & 0.803 & 0.644 & 0.745 & 0.877 & \textbf{0.812} & 0.859 & 0.856 & 0.641 & 0.789 & 0.730 & 0.706 \\
${\footnotesize \sf{Proposed}}_1$ & \textbf{0.743} & \textbf{0.639} & 0.686 & 0.402 & \textbf{0.862} & \textbf{0.805} & \textbf{0.755} & 0.889 & 0.805 & \textbf{0.891} & \textbf{0.935} & 0.639 & 0.579 & \textbf{0.850} & 0.741 \\
${\footnotesize \sf{Proposed}}_2$ & 0.734 & 0.609 & 0.688 & 0.388 & 0.837 & 0.790 & 0.740 & \textbf{0.896} & 0.786 & 0.882 & 0.926 & 0.639 & 0.570 & 0.845 & 0.718 \\ \hline \hline
 & \multicolumn{2}{c|}{TID2013} & \multicolumn{5}{c|}{CSIQ} & \multicolumn{6}{c|}{LIVE} & \multicolumn{2}{c}{\multirow{2}{*}{\centering Avg.}} \\
 & \#24 & \dbcol{ALL} & WN & JPEG & JP2k & Gblur & \dbcol{ALL} & JP2k & JPEG & WN & Gblur & FF & \dbcol{ALL}  \\ \hline
BRISQUE-L & 0.830 & \dbcol{0.762} & 0.688 & 0.786 & 0.804 & 0.834 & \dbcol{0.781} & 0.885 & 0.939 & 0.960 & 0.891 & 0.862 & \dbcol{0.917} & \multicolumn{2}{c}{0.802} \\
BLIINDS-II & 0.795 & \dbcol{0.651} & 0.755 & 0.823 & 0.798 & 0.817 & \dbcol{0.804} & 0.932 & 0.933 & 0.930 & 0.921 & 0.880 & \dbcol{0.923} & \multicolumn{2}{c}{0.743} \\
${{\mathbf f}}^{SRN}{\footnotesize \sf{SVR}}$  & 0.903 & \dbcol{0.715} & 0.887 & 0.900 & \textbf{0.911} & \textbf{0.897} & \dbcol{\textbf{0.908}} & 0.930 & 0.953 & 0.967 & 0.932 & 0.840 & \dbcol{0.938} & \multicolumn{2}{c}{0.802} \\
${{\mathbf f}}^{all}{\footnotesize \sf{SVR}}$ & 0.864 & \dbcol{0.753} & \textbf{0.933} & \textbf{0.908} & 0.894 & 0.886 & \dbcol{0.903} & 0.922 & \textbf{0.957} & \textbf{0.978} & 0.933 & \textbf{0.895} & \dbcol{\textbf{0.947}} & \multicolumn{2}{c}{0.825} \\
CORNIA & 0.884 & \dbcol{\textbf{0.777}} & 0.749 & 0.878 & 0.886 & 0.872 & \dbcol{0.867} & 0.873 & 0.919 & 0.939 & 0.938 & 0.888 & \dbcol{0.915} & \multicolumn{2}{c}{\textbf{0.826}} \\
${\footnotesize \sf{Proposed}}_1$ & \textbf{0.908} & \dbcol{0.753} & 0.787 & 0.849 & 0.849 & 0.844 & \dbcol{0.830} & \textbf{0.935} & 0.943 & 0.958 & \textbf{0.942} & 0.862 & \dbcol{0.923} & \multicolumn{2}{c}{0.807} \\
${\footnotesize \sf{Proposed}}_2$ & 0.901 & \dbcol{0.767} & 0.891 & 0.859 & 0.854 & 0.841 & \dbcol{0.855} & 0.927 & 0.949 & 0.961 & \textbf{0.942} & 0.862 & \dbcol{0.927} & \multicolumn{2}{c}{0.820} \\ \hline \hline
\end{tabular}
\setcounter{table}{\value{MYtemptabcnt}}
\end{table*}

For comparison purposes, we linearly mapped the TID2013 MOS scores and the CSIQ database DMOS scores to the range of the LIVE database DMOS scores, respectively. State-of-the-art BIQA models were then learned on the training images based on the realigned quality scores and then applied to the test images. In addition, we randomly extracted 3000 PIPs from all the training images based on the realigned quality scores with $T=10$, and trained the proposed approach on them. Similarly, the train-test procedure was repeated 100 times and the median SRCC values were reported in Table \ref{ExtensionHybrid}. The proposed approach trained in the first case is referred to as ${\small \sf{Proposed}}_1$, and that in the second case is referred to as ${\small \sf{Proposed}}_2$ in Table \ref{ExtensionHybrid}.

As can be seen, both ${\small \sf{Proposed}}_1$ and ${\small \sf{Proposed}}_2$ are highly comparable to state-of-the-art BIQA methods across all three databases. In particular, they outperform the other algorithms for most distortion categories in the TID2013, i.e. \# 1, 2, 5, 6, 8-10, 22, and 24. On the CSIQ and LIVE database, they perform better than BRISQUE-L and BLIINDS but are inferior to ${{\mathbf f}}^{SRN}{\small \sf{SVR}}$, ${{\mathbf f}}^{all}{\small \sf{SVR}}$, and CORNIA, indicating that there is still scope to improve performance. It is important to note that better performance can be obtained by adding more PIPs to the training data.

${\small \sf{Proposed}}_2$ yields better performance than ${\small \sf{Proposed}}_1$, because the information about the relative quality between images from different databases is not included in the training set of ${\small \sf{Proposed}}_1$, . This is similar to the previous experiment on the LIVE database; thus, to apply the proposed BIQA framework to emerging distortion categories or to extend existing PIP databases with new image pairs, it would better to add PIPs composed of both new and existing images besides those composed purely of new images.

Although ${{\mathbf f}}^{all}{\small \sf{SVR}}$ and CORNIA obtain better performance by using the realigned quality scores, it should be noted that the realignment is not precise and relies heavily on the hypothesis that the images included in each database cover the full range of quality with small intervals. Thus for the extension of these BIQA approaches, it is necessary to collect sufficient images associated with emerging kinds of distortion at diverse levels of degradation and to recruit many observers to evaluate their quality scores. As has been discussed, the collection of such images and the acquisition of the quality scores is difficult and inconvenient, leading to a huge cost to extend these algorithms.

In contrast, we can generate PIPs with an arbitrary number of images afflicted by the new distortion category at very low cost, and then simply add them to the existing PIP dataset to learn a robust BIQA model. Moreover, only a few subjects are needed to label the relative quality of each PIP; thus it is easy and efficient to extend the proposed framework, and our method can predict perceptual quality scores in high consistency with human perception, as verified in both Tables \ref{ExtensionLIVE} and \ref{ExtensionHybrid}.

\section{Subjective Study PIPs}
\label{Subjective Study PIPs}

In this section, we present an extensive subjective study of PIPs through a paired-comparison experiment. We adopted the 808 images (29 reference images and 779 distorted ones) contained in the LIVE database to construct image pairs. In total, there are 326,028 ($=808\times 807/2$) possible pairs, from which we randomly selected approximately 240,000 pairs for evaluation in accordance with \cite{bt2002500}. We conducted such a large number of comparisons with the objective of statistically investigating the relationship between subjective quality scores and preference labels, and validating the efficiency and efficacy of our recommended PIP-generation methods in Section \ref{sec:Generation of Preference Image Pairs}.

We randomly divided all the pairs into 10 portions and appointed one subject for each portion. The subjects for the study are students between 20 and 30 years old at Xidian University. Most image pairs were evaluated by only one subject each. In addition, 10,000 image pairs were evaluated by all the subjects. In addition, we have approximately 240,000 pairs evaluated by each a single subject. The average preference labels for the 10,000 pairs labeled by all the subjects is regarded as the benchmark for evaluating the validity of the raw labels assigned by each subject.

In the experiment, subjects were shown the prepared image pairs in random order and could divide image pairs into an arbitrary number of sessions. The maximum time of each session was limited to 30 minutes to minimize the effect of observer fatigue \cite{bt2002500}. The experimental setup we used was a double-stimulus methodology in which both of the images in a certain pair were simultaneously shown on a monitor displaying at resolution of $1680\times 1080$ pixels. Each image was displayed at its original resolution. The test framework was developed in MATLAB. The procedures of the experiments and the processing of the preference labels are the same as presented in Section III.A.

The evaluation of an image pair in our experiment cost 3 seconds on average, demonstrating the efficiency of generating PIPs. Analysis of the obtained data is detailed in the following.

\subsection{Relationship between Subjective Preference Labels and Quality Scores}
\label{sec:Relationship}

\begin{figure}[t]
\centering
\includegraphics[width=3.5in]{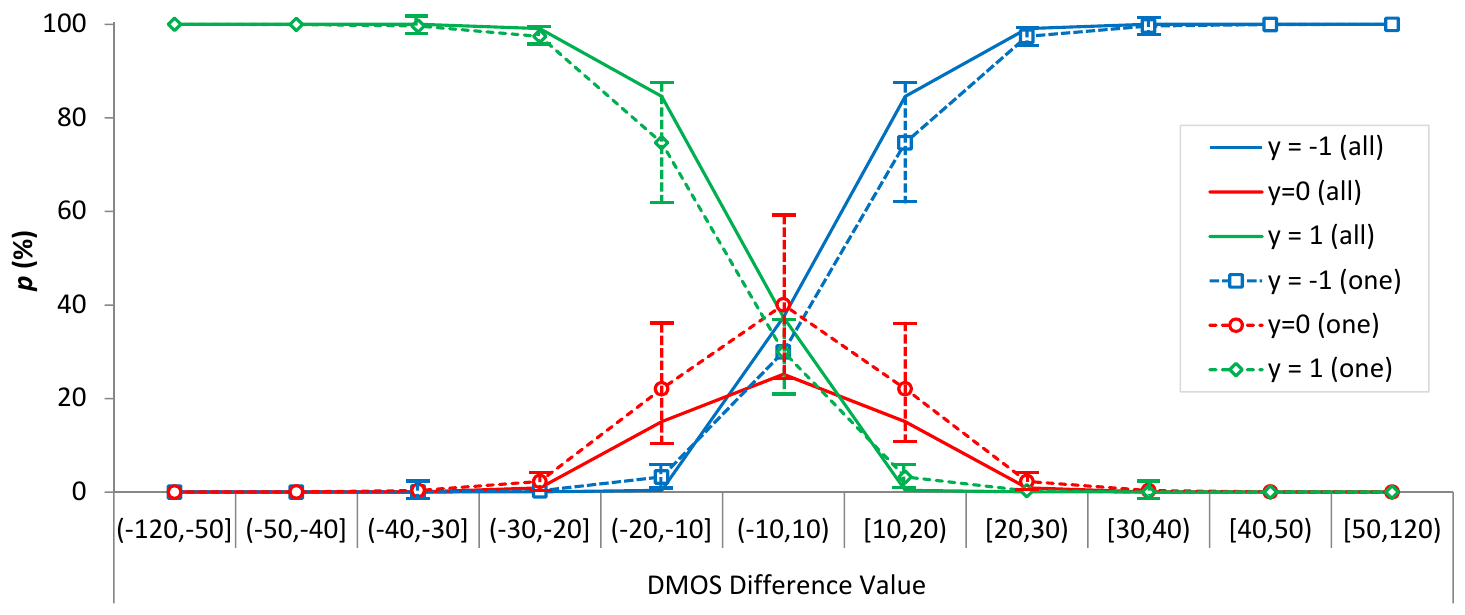}
\caption{The relation between the DMOS difference value and the preference label.}
\label{SubDataAnalysis}
\end{figure}

To visualize the relationship between the subjective preference label and the quality score, we plot the distributions of the preference label versus the DMOS difference for the 10,000 pairs labeled by all the subjects (marked by "all" in Fig. \ref{SubDataAnalysis}). In addition, we calculated the distributions for the pairs labeled by each subject, respectively, and show the average distribution (shown in the form of dashed lines and marked by "one" in Fig. \ref{SubDataAnalysis}) and the standard deviations of these distributions across subjects (shown in the form of error-bars). The x-coordinate shows the interval of the DMOS difference, and the y-coordinate is the percentage of corresponding pairs assigned as each kind of preference label.

The distributions of the singly assigned labels approach to those of the mean responses of a number of observers. In particular, when the DMOS difference is small, the variances on the distributions are relatively large. This is mainly because subjects label different numbers of image pairs as "uncertain" (i.e. $y=0$). In contrast, when the DMOS difference is sufficiently large, the variances become very small, because subjects usually report unanimous preference labels. Thus we can conclude that the distributions are generally consistent across subjects.

In Table \ref{tab:ACCURACIES1}, we report the median accuracy of the responses by each subject and the accuracy of the preference labels derived from DMOS for the 10,000 pairs. In Table \ref{tab:ACCURACIES2}, we tabulate the accuracies without considering the pairs associated with a preference label of $y=0$. As is clear, when the DMOS difference is small, DMOS cannot precisely reflect the relative quality of the corresponding images; but if the DMOS difference is sufficiently large, e.g. greater than 10, it is generally consistent with the human judgment of the relative quality. This agrees with the impact of the threshold $T$ on the performance of the proposed BIQA method, as discussed in Section V.A.

\begin{table}[b]
\centering
\setcounter{MYtemptabcnt}{\value{table}}
\setcounter{table}{8}
\caption{Accuracies of The Responses from Each Subject and Preference Labels Derived from DMOS. (\%) }
\label{tab:ACCURACIES1}
\setlength{\tabcolsep}{0.08in}
\begin{tabular}{l|cccccc}\hline \hline
	    & \multicolumn{6}{c}{intervals of absolute value of the DMOS difference} \\
	    & [0,10) & [10,20)	& [20,30)	& [30,40)	& [40,50)	& [50,120) \\ \hline
Subject	& 85.58	 & 92.23	& 99.01	    & 99.90	    & 100	    & 100 \\
DMOS	& 88.14	 & 94.02	& 99.27	    & 100	    & 100	    & 100 \\  \hline \hline
\end{tabular}
\setcounter{table}{\value{MYtemptabcnt}}
\end{table}

\begin{table}[b]
\centering
\setcounter{MYtemptabcnt}{\value{table}}
\setcounter{table}{9}
\caption{The Accuracies Without Considering the Pairs Associated With a Preference Label of $y=0$. (\%)}
\label{tab:ACCURACIES2}
\setlength{\tabcolsep}{0.08in}
\begin{tabular}{l|cccccc}\hline \hline
	    & \multicolumn{6}{c}{intervals of absolute value of the DMOS difference} \\
	    & [0,10) & [10,20)	& [20,30)	& [30,40)	& [40,50)	& [50,120) \\ \hline
Subject	& 96.18	& 98.29	& 100	& 100	& 100	& 100 \\
DMOS	& 98.39	& 99.61	& 100	& 100	& 100	& 100 \\  \hline \hline
\end{tabular}
\setcounter{table}{\value{MYtemptabcnt}}
\end{table}

It is remarkable that the judgments made by a single subject yield a highly comparable level of accuracy obtained by the DMOS data labelled by a number of subjects. Thus, the proposed scheme significantly saves the cost for collecting training data. These results demonstrate that we can obtain valid preference labels by making only one subjective judgment for each image pair. Compared with the acquisition of DMOS, this is much easier and more efficient; we can learn an efficient BIQA model from the labeled image pairs, as shown in Section IV and the following subsection.

\subsection{IQA Performance}
\label{sec:IQA Performance}

In order to further test the validity of the obtained subjective preference labels, we conducted IQA experiments on the LIVE database similarly as introduced in Section V.B. Specially, we randomly selected 20 groups of images from the LIVE database and 2,000 PIPs from the corresponding subjectively labeled image pairs for training. Afterward, we applied the learned model to the test images to predict the corresponding quality scores. For the purpose of comparison, we also constructed training set based on the DMOS. The median KRCC, SRCC, and PLCC values on the entire LIVE database were reported in Table \ref{tab:PfmSubject}.

In addition, we calculated the difference feature vectors of the subjectively labeled image pairs which are purely composed of the test images and then fed them into the learned classifier to estimate the corresponding preference labels. We computed four accuracies to evaluate the prediction precision. The first one is the accuracy on all the corresponding test image pairs and is denoted as ${Acc}^{All}$. The second one is operated only on the pairs of which the difference DMOS is consistent with the subjective judgment, and is denoted as ${Acc}^{cnst}$. Moreover, we estimated the corresponding preference labels by comparing the predicted quality scores, and denote the corresponding preference accuracies ${Acc}^{All}_Q$ and ${Acc}^{cnst}_Q$.

\begin{table}[t]
\centering
\setcounter{MYtemptabcnt}{\value{table}}
\setcounter{table}{10}
\caption{Performance of the Proposed BIQA Method on the Entire LIVE Database.}
\label{tab:PfmSubject}
\setlength{\tabcolsep}{0.05in}
\begin{tabular}{l|ccccccc}\hline \hline
 & KRCC & SRCC & PLCC & ${Acc}^{All}$ & ${Acc}^{cnst}$ & ${Acc}^{All}_Q$ & ${Acc}^{cnst}_Q$ \\ \hline
DMOS & 0.783 & 0.935 & 0.938 & 0.949 & 0.957 & 0.956 & 0.963 \\
Subject & 0.783 & 0.935 & 0.937 & 0.948 & 0.956 & 0.956 & 0.963 \\ \hline \hline
\end{tabular}
\setcounter{table}{\value{MYtemptabcnt}}
\end{table}

It is obvious that the proposed BIQA method obtains almost the same performance by adopting the subjective preference labels or the labels derived from DMOS, demonstrating the validity of the preference labels obtained through paired comparisons. It is also notable that the performance here is similar to that reported in Section V.B. This corroborates the expectation that by choosing a proper threshold, we can generate valid PIPs from existing IQA databases for learning a robust BIQA model. In addition, ${Acc}^{All}_Q$ and ${Acc}^{cnst}_Q$ are slightly greater than ${Acc}^{All}$ and ${Acc}^{cnst}$. The reason is that more comparisons are performed on each test image in the quality prediction procedure, indicating the efficacy of the proposed quality prediction approach.

\section{Conclusion}
\label{sec:Conclusion}
In this paper, we presented a BIQA framework that learns to predict perceptual image quality scores from preference image pairs. Thorough experiments on the four largest standard databases demonstrate that our method correlates highly with human perceptions of quality and that the framework can be easily extended to emerging distortions. In addition, an extensive subjective study corroborates the efficiency and validity of generating PIPs through paired comparisons or from existing IQA databases. In our continued search for convenient but effective PIP-generation methods and BIQA metrics, we have deployed the simplest operations in both the subjective study and the proposed BIQA approach in this paper. There is still room for the optimization of the subjective study and the improvement of IQA performance. In addition, the low cost of the generation of PIPs, coupled with the efficacy of the proposed BIQA approach, make the deployment of this framework to solve other quality assessment (QA) problems, e.g. other types of images \cite{Wang2014IJCV}\cite{Wang2013TNNLS}, esthetic assessment \cite{Li2014Esthetic}, performance evaluation \cite{Li2014MAP}, etc., a promising option. Further research is needed to explore quality-relevant features \cite{Li2014feature}\cite{Zhou2014feature} and generate corresponding stimulus pairs with valid preference labels.

\begin{IEEEbiography}[{\includegraphics[width=1in,height=1.25in,clip,keepaspectratio]{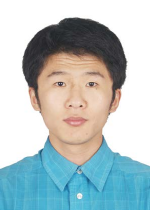}}]{Fei Gao}
Fei Gao received the B.Sc. degree in electrical information engineering from Xidian University in 2009. Now he is pursuing the Ph.D. degree in intelligent information processing with the VIPS Laboratory, School of Electronic Engineering, Xidian University. From October 2012 to October 2013, he studied at the University of Technology, Sydney, NSW, Australia as a visiting Ph.D. student. His research interests include computer vision and machine learning.
\end{IEEEbiography}

\begin{IEEEbiography}[{\includegraphics[width=1in,height=1.25in,clip,keepaspectratio]{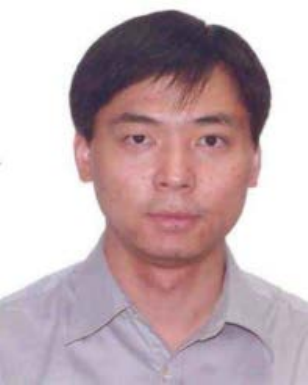}}]{Dacheng Tao}
Dacheng Tao (M'07-SM'12) is Professor of Computer Science with the Centre for Quantum Computation \& Intelligent Systems, and the Faculty of Engineering and Information Technology in the University of Technology, Sydney. He mainly applies statistics and mathematics for data analytics and his research interests spread across computer vision, data mining, geoinformatics, image processing, machine learning, multimedia and video surveillance. His research results have expounded in one monograph and 100+ publications at prestigious journals and prominent conferences, such as IEEE T-PAMI, T-NNLS, T-IP, T-SP, T-KDE, T-CYB, JMLR, IJCV, NIPS, ICML, CVPR, ICCV, ECCV, AISTATS, ICDM; ACM SIGKDD and Multimedia, with several best paper awards, such as the best theory/algorithm paper runner up award in IEEE ICDM'07 and the best student paper award in IEEE ICDM'13.
\end{IEEEbiography}

\begin{IEEEbiography}[{\includegraphics[width=1in,height=1.25in,clip,keepaspectratio]{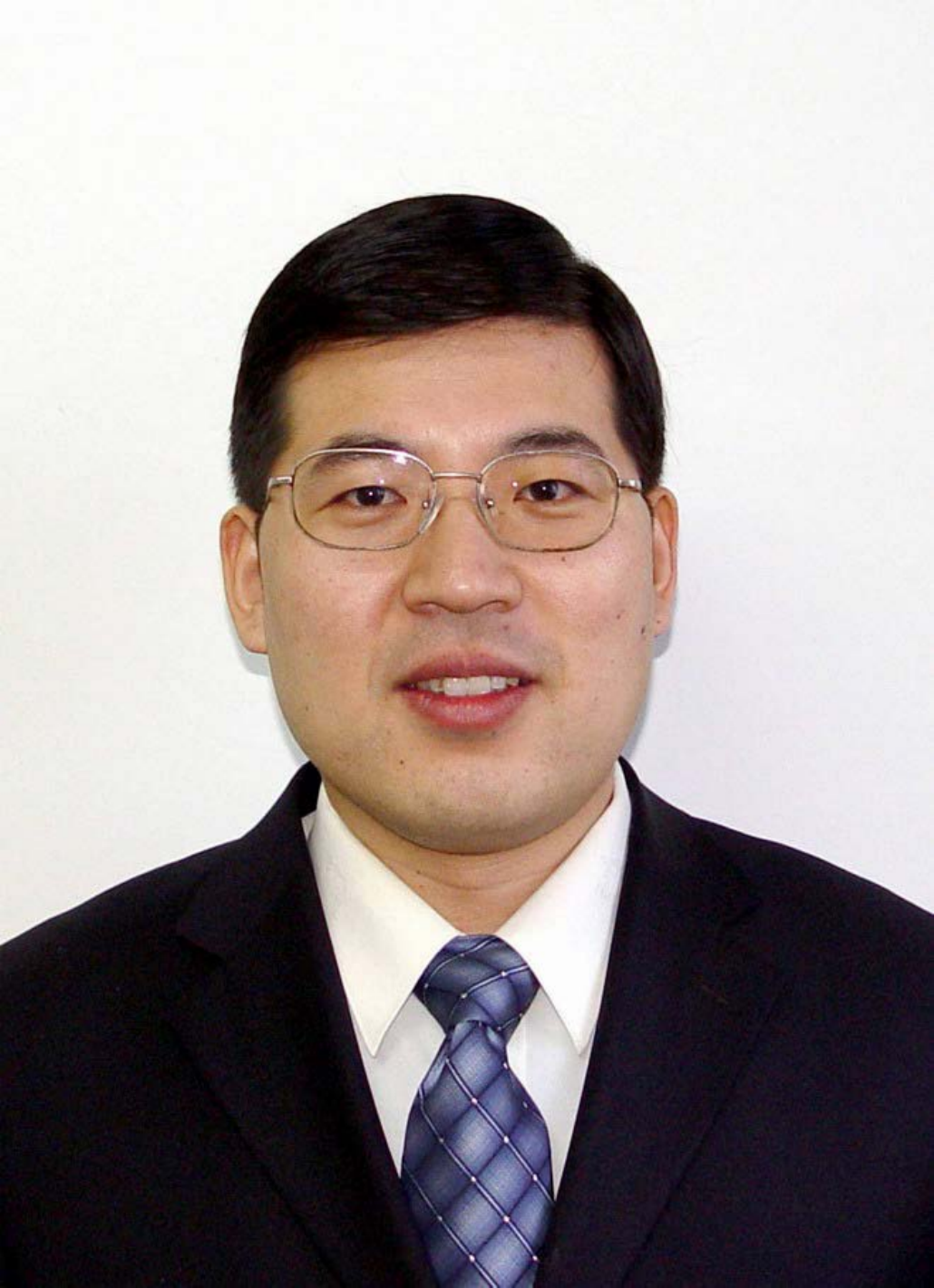}}]{Xinbo Gao}
Xinbo Gao (M'02-SM'07) received the B.Eng., M.Sc. and Ph.D. degrees in signal and information processing from Xidian University, China, in 1994, 1997 and 1999 respectively. From 1997 to 1998, he was a research fellow in the Department of Computer Science at Shizuoka University, Japan. From 2000 to 2001, he was a postdoctoral research fellow in the Department of Information Engineering at the Chinese University of Hong Kong. Since 2001, he joined the School of Electronic Engineering at Xidian University. Currently, he is a Cheung Kong Professor of Ministry of Education, China, a Professor of Pattern Recognition and Intelligent System, and Director of the State Key Laboratory of Integrated Services Networks. His research interests are computational intelligence, machine learning, computer vision, pattern recognition and wireless communications. In these areas, he has published 5 books and around 200 technical articles in refereed journals and proceedings including IEEE TIP, TCSVT, TNN, TSMC, IJCV, Pattern Recognition etc.. He is on the editorial boards of several journals including Signal Processing (Elsevier), and Neurocomputing (Elsevier). He served as general chair/co-chair or program committee chair/co-chair or PC member for around 30 major international conferences. Now, he is a Fellow of IET and Senior Member of IEEE.
\end{IEEEbiography}

\begin{IEEEbiographynophoto}{Xuelong Li}
Xuelong Li (M'02-SM'07-F'12) is a full professor with the Center for OPTical IMagery Analysis and Learning (OPTIMAL), State Key Laboratory of Transient Optics and Photonics, Xi'an Institute of Optics and Precision Mechanics, Chinese Academy of Sciences, Xi'an 710119, Shaanxi, P. R. China.
\end{IEEEbiographynophoto}

\vfill





\end{document}